%% file: article.tex
\documentclass[dvipsnames,format=sigconf]{acmart}
\AtBeginDocument{%
\providecommand\BibTeX{{%
\normalfont B\kern-0.5em{\scshape i\kern-0.25em b}\kern-0.8em\TeX}}}

\usepackage[]{algorithm}
\usepackage{algpseudocode}
\algnewcommand\And{\textbf{and}}
\usepackage{amsfonts}
\usepackage{caption}
\usepackage{rotating}
\usepackage{xcolor}
\usepackage{multirow}
\usepackage{color}
\usepackage{array}
\usepackage{booktabs}
\usepackage{tabularx}
\usepackage{multirow}
\usepackage{placeins}
\usepackage{longtable}
\usepackage{supertabular,enumitem}
\usepackage{float}
\usepackage{hyperref}
\usepackage{academicons}
\definecolor{orcidlogocol}{HTML}{A6CE39}
\usepackage{xspace}
\usepackage{booktabs}
\usepackage{tikz,pgfplots}
\usepackage{amsmath}

\usetikzlibrary{arrows.meta}
\usepackage{eurosym}
\DeclareUnicodeCharacter{20AC}{\EUR{}}

\usepackage{diagbox}

\usepackage[caption=false]{subfig}

\newcolumntype{L}[1]{>{\raggedright\let\newline\\\arraybackslash\hspace{0pt}}m{#1}}

\xspace
\newcommand{\TITLE}{Robust Multi-Objective Optimization for Bicycle Rebalancing in Shared Mobility Systems\xspace}

\newcommand{\ktrucks}{$\boldsymbol{\vec{t}}$\xspace}
\newcommand{\routes}{$\mathcal{\mathbf{R}}$\xspace}

\newcommand{\ABtwo}{Random Station Rebalancing Between Routes\xspace}

\newcommand{\BBoneMax}{Deterministic Station Rebalancing with Maximum Difference\xspace}
\newcommand{\BBone}{Deterministic Station Rebalancing with Minimum/Maximum Difference\xspace}
\newcommand{\BBtwo}{Station Rebalancing using Roulette-Wheel Selection\xspace}

\copyrightyear{2026}
\acmYear{2026}
\setcopyright{acmlicensed}\acmConference[GECCO '26]{Genetic and
Evolutionary Computation Conference}{July 13-17, 2026}{San Jos\'e, Costa Rica}
\acmBooktitle{Genetic and Evolutionary Computation Conference (GECCO '26),
July 13-17, 2026}
\acmDOI{XXXXXXX}
\acmISBN{XXXXXX}

\begin{document}

\title{\TITLE}

\author{Diego Daniel Pedroza-Perez}
\authornote{Corresponding authors.}
\authornote{All authors contributed equally to this research.}
\orcid{0009-0008-8513-1427}
\affiliation{%
 \institution{ITIS Software, University of Malaga, Spain}
 \country{}
}
\email{pedroza@uma.es}

\author{Gabriel Luque}
\orcid{0000-0001-7909-1416}
\affiliation{%
 \institution{ITIS Software, University of Malaga, Spain}
 \country{}
}
\email{gluque@uma.es}

\author{Sergio Nesmachnow}
\orcid{0000-0002-8146-4012}
\affiliation{%
    \institution{Universidad de la República, Uruguay}
    \country{}
}
\email{sergion@fing.edu.uy}

\author{Jamal Toutouh}
\orcid{0000-0003-1152-0346}
\affiliation{%
 \institution{ITIS Software, University of Malaga, Spain}
 \country{}
}
\email{jamal@uma.es}



\begin{abstract}

Dock-based bike-sharing systems exhibit spatial imbalances between bicycle supply and user demand, often addressed through overnight truck-based rebalancing. This work studies static overnight rebalancing under demand uncertainty modeled as a tri-objective optimization problem. The objectives minimize total travel distance, expected unmet demand, and a robustness-oriented unmet demand measure over high-demand scenarios.
Route plans are evaluated via a recourse simulation that enforces truck loads and station capacity constraints across multiple demand realizations. The robustness objective supports selecting plans that reduce peak-demand service degradation. Trade-off solutions are approximated with Non-dominated Sorting Genetic Algorithm II using a permutation--partition encoding and domain-specific relocation operators, including a biased best-improvement move for station relocation.
Experiments on the real Barcelona Bicing system with 460 stations show well-distributed Pareto sets and substantial contributions to the reference non-dominated set. Greedy constructive baselines mainly yield extreme solutions and are often dominated.

\end{abstract}


\setlength{\textfloatsep}{6pt plus 2pt minus 2pt}   


\maketitle


\section{Introduction}
Dock-based bike-sharing systems play an essential role in sustainable urban mobility strategies. A recurring operational challenge in these systems is spatial imbalance: user movements can exhaust bicycles at some stations while saturating docks at others, reduced service quality and user satisfaction \citep{SHUI2020102648}. Operators address this through overnight truck-based rebalancing, where a static plan specifies routes, pickup and delivery quantities, and enforces vehicle capacity constraints. The problem is interdependent since early pickups affect the feasibility of later deliveries \citep{RAVIV2013}.

Bicycle and dock demand varies across stations and time periods, involving uncertainty that forecasting cannot fully capture. Deterministic plans may perform well under typical conditions but fail under atypical demand \citep{DELLAMICO2018362}. A scenario-based framework offers greater realism by evaluating rebalancing plans across multiple demand realizations from historical data.
%

The static rebalancing task is modeled as a vehicle routing problem (VRP) with load-dependent feasibility constraints and service-oriented objectives. Given the large number of stations and stochastic demand in real systems, optimization methods must efficiently approximate the Pareto front across conflicting goals. Existing studies propose exact, heuristic, and metaheuristic approaches for various problem variants~\cite{POULIASI2025100141, su13041829, Neumann-Saavedra2025}, including multiobjective models balancing operational effort and service quality~\cite{JIABEE2020, LI2021102216}.


This study addresses the overnight static rebalancing problem in large-scale dock-based bike-sharing systems under scenario-based demand uncertainty. A tri-objective model is proposed to minimize travel distance, expected unmet demand, and performance degradation under high-demand conditions. 
To explore trade-offs among the three objectives, a multiobjective evolutionary algorithm (MOEA), the Non-dominated Sorting Genetic Algorithm II (NSGA-II) \citep{Deb2001}, is employed. NSGA-II is a well-established method in multi-objective optimization, widely recognized for its accuracy~\citep{nesmachnowX2018comparison, toutouh2020soft, CintranoJamal2022stations}. The proposed method includes: (i) a permutation–partition encoding for route sequencing and segmentation, (ii) relocate-based mutation operators for inter-route station transfers, and (iii) a biased relocation operator, that balances exploratory diversity with route efficiency. Computational experiments on the Barcelona Bicing system (460 stations) evaluate algorithmic performance and analyze representative trade-offs through spatial summaries.

\sloppy
The main contributions of this paper are: (a) a tri-objective formulation of static overnight bike rebalancing that specifically considers robustness as an objective defined as unmet demand over high-demand scenarios; (b) a permutation--partition encoding for multi-truck route sets with an explicit unvisited-station segment that supports permutation operators without feasibility repair; (c) domain-specific inter-route relocate mutation operators, including a deterministic best-improvement diversification variant and a roulette-biased stochastic variant; and (d) an empirical evaluation on the real-world Barcelona Bicing scenario that compares mutation operators against greedy baselines and an ablated variant.

The article is organized as follows. Section~\ref{sec:problem} describes the problem formulation. Section~\ref{sec:algorithms} describes the proposed approach. Later, Section~\ref{sec:experimental-setup} describes the case study data and experimental design, and Section~\ref{sec:experimental-analysis} reports the experimental analysis. Section~\ref{sec:conclusions} presents the conclusions and formulates the main lines for future work.

\input{problem_definition_jamal}


\section{Method}
\label{sec:algorithms}
This section describes the proposed solution approach for the tri-objective rebalancing problem. The optimization backbone is a parallel MOEA (pMOEA) based on standard NSGA-II~\cite{Deb2001}. The remainder of the section focuses on the problem-specific components: the permutation--partition encoding (\routes,\ktrucks), the standard permutation operators applied to \routes, and the proposed domain-specific relocation mutations that exploit structural properties of the problem to guide the search toward high-quality trade-offs.

\subsection{Operators and parallel implementation}
\label{sec:operators}

\paragraph{Initialization and Solution encoding.} 
The initial population is generated by random initialization. Each solution is encoded by two integer vectors: \routes with \(|\mathcal{S}|\) elements and \ktrucks with \(|\mathcal{T}|+1\) entries. \routes encodes a permutation of the stations, determining their visiting order. 
The vector \ktrucks partitions \routes into \(|\mathcal{T}|\) routes, one for each truck, and one additional segment that collects unvisited stations. 

\begin{figure}[!ht]
\centering
\includegraphics[width=0.8\linewidth]{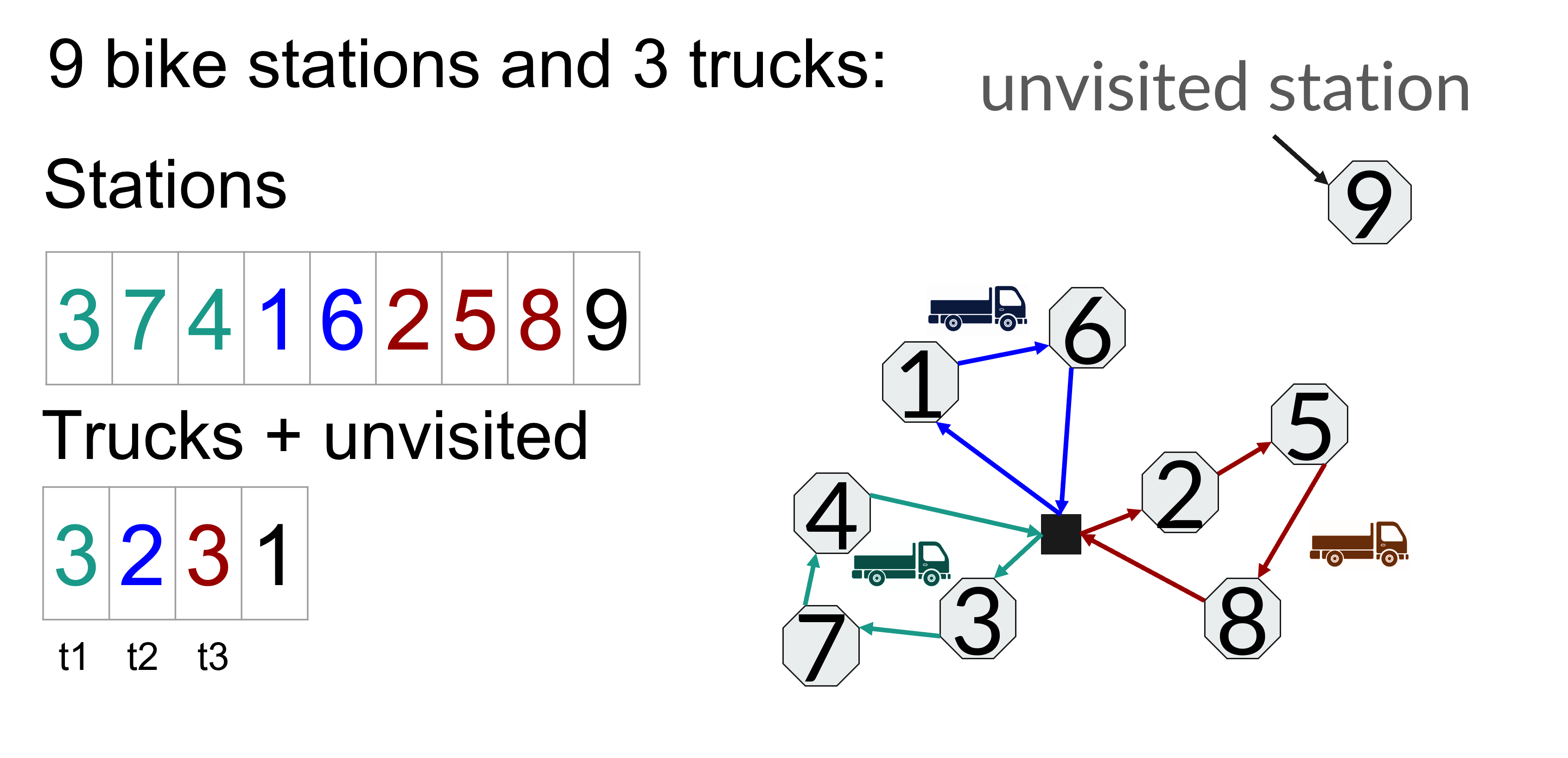}
\caption{Example of solution encoding of a scenario with 9 stations and 3 trucks.}
  \label{fig:solution-encoding}
\end{figure}

\figurename~\ref{fig:solution-encoding} illustrates the encoding. In the example, encoding \routes and \ktrucks yields $\mathbf{R}=\{R_t\}_{t\in\mathcal{T}}$: stations 3, 7, and 4 to truck 1; stations 1 and 6 to truck 2; and stations 2, 5, and 8 to truck 3; with remaining stations unvisited. This encoding enables permutation-based crossovers/mutations on \routes and partition operators on \ktrucks, avoiding feasibility repairs since \routes is a permutation. Relocation moves transfer a station $s$ between truck segments in (\routes,\ktrucks), updating \ktrucks while preserving the partition structure---equivalent to classical VRP 1--0 relocate without repair. 

\paragraph{Selection, replacement, and fitness assignment.} 
NSGA-II uses the ($\mu$+$\lambda$) evolution model. The selection process is based on dominance through crowding distance selection for NSGA-II. Fitness assignment is performed by considering Pareto dominance rank, valid results, and crowding distance value.

\paragraph{Parallel master-slave implementation}
The parallel NSGA-II proposed in this work aligns with the master-slave model~\cite{alba2013parallel}.
The pMOEA proposed in this study is structured hierarchically into a master-slave framework, with a master process responsible for evolutionary search and oversight of a group of slave processes that compute the fitness function evaluations.

\paragraph{Standard recombination and mutation operators.} 
The applied crossover operators are standard permutation methods selected for their complementarity strengths. Order Crossover (OX) performs especially well in permutation-based problems involving graphs or routes. Likewise, Edge Recombination Crossover (ERX) and Enhanced Edge Recombination Crossover (EERX) preserve adjacency information between edges, thereby maintaining essential structural relationships in the offspring. Although Partially Mapped Crossover (PMX) is less effective in retaining edge information, it preserves the relative order of elements, which suits the characteristics of the problem~\cite{surveyOperatorsPermutations}. 


Four standard mutation operators that modify the route permutation are applied in this paper: 
Block-Move (BMM), which relocates a contiguous block of $n$ elements to another position; 
Block-Swap (BSM), which exchanges two contiguous blocks of $n$ elements;
Swap (SM), a special case of BSM with $n=1$;
and Inversion (IM), which reverses the subsequence between two selected positions.



\subsection{Domain-specific mutation operators}\label{subsection:ktrucks_Mut_Oper}
This work defines domain-specific mutation operators that modify the route partition by relocating a single station between two routes with probability $PM_T$. Mutations act on the encoding $($\routes,\ktrucks$)$ and preserve the station permutation in \routes, except for the relocated element. Each station is assigned to at most one truck route or remains unvisited. The operators differ in the rule used to select the relocated station and its insertion position.

\begin{figure}[!h]
\setlength{\abovecaptionskip}{0pt}
\setlength{\belowcaptionskip}{0pt}
\includegraphics[width=1\linewidth]{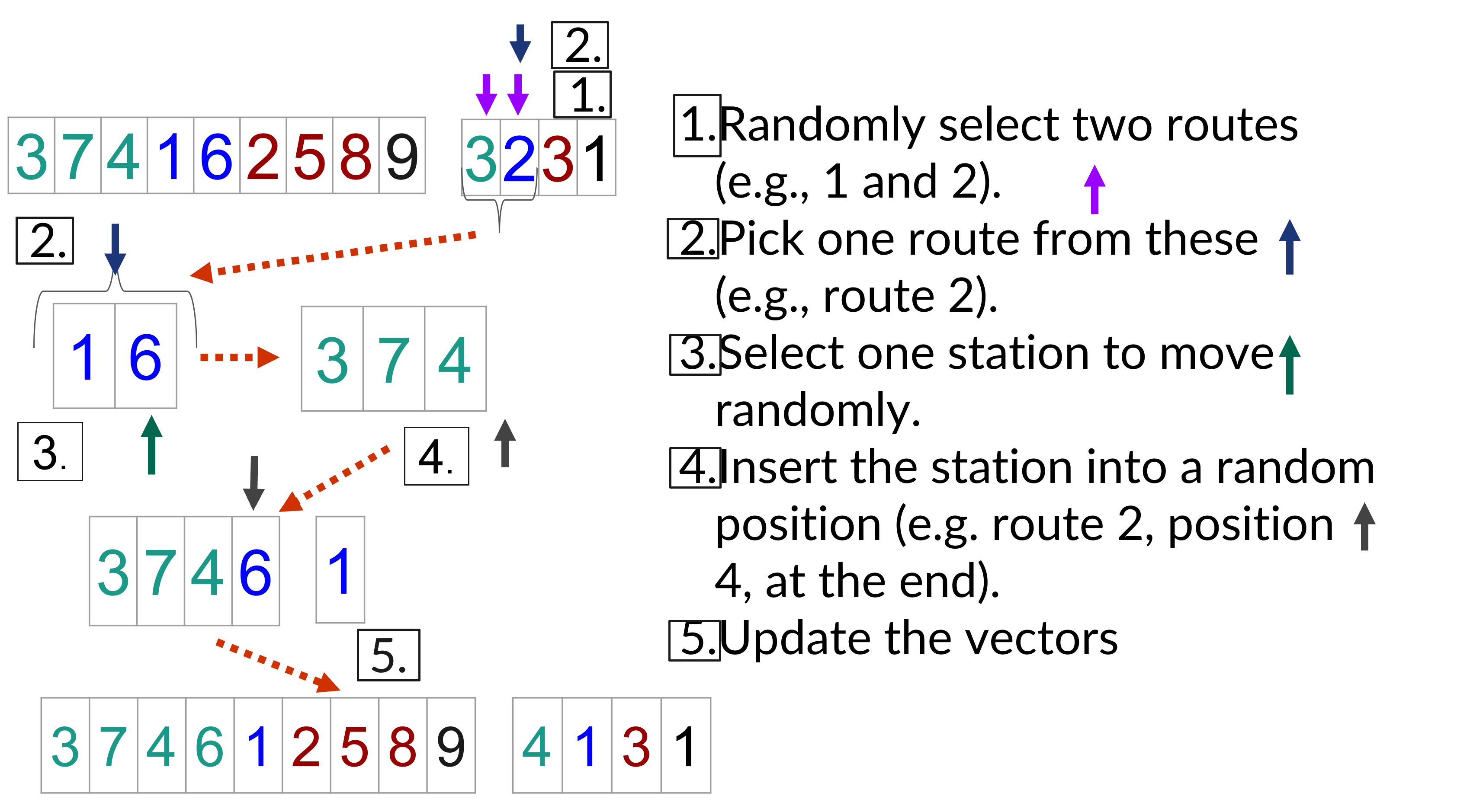}
\caption{Diagram showing the operation of the AB2 operator. The red arrows indicate the step sequence.}
\label{mut:AB2}
\end{figure}

\paragraph{Random relocation.}
The \ABtwo (AB2) operator implements an unbiased 1--0 relocate move between two routes.
The operator randomly selects two distinct routes, designates one as the source and the other as the destination, samples one station from the source route, and inserts it at a uniformly random position in the destination route.
Algorithm~\ref{algo:AB2} and \figurename~\ref{mut:AB2} summarize the procedure.

\begin{algorithm}[!t]
\small
\caption{\ABtwo}\label{algo:AB2}
\begin{algorithmic}[1]
\State Randomly select two distinct routes.
\State Select one route as the source and set the other as the destination.
\State Randomly choose a station from the source route.
\State Insert the station into a random  position in the destination route.
\State Update the solution encoding (\routes,\ktrucks) accordingly.
\end{algorithmic}
\end{algorithm}

\begin{figure}[!h]
\includegraphics[trim=0 5mm 0 5mm, clip, width=1\linewidth]{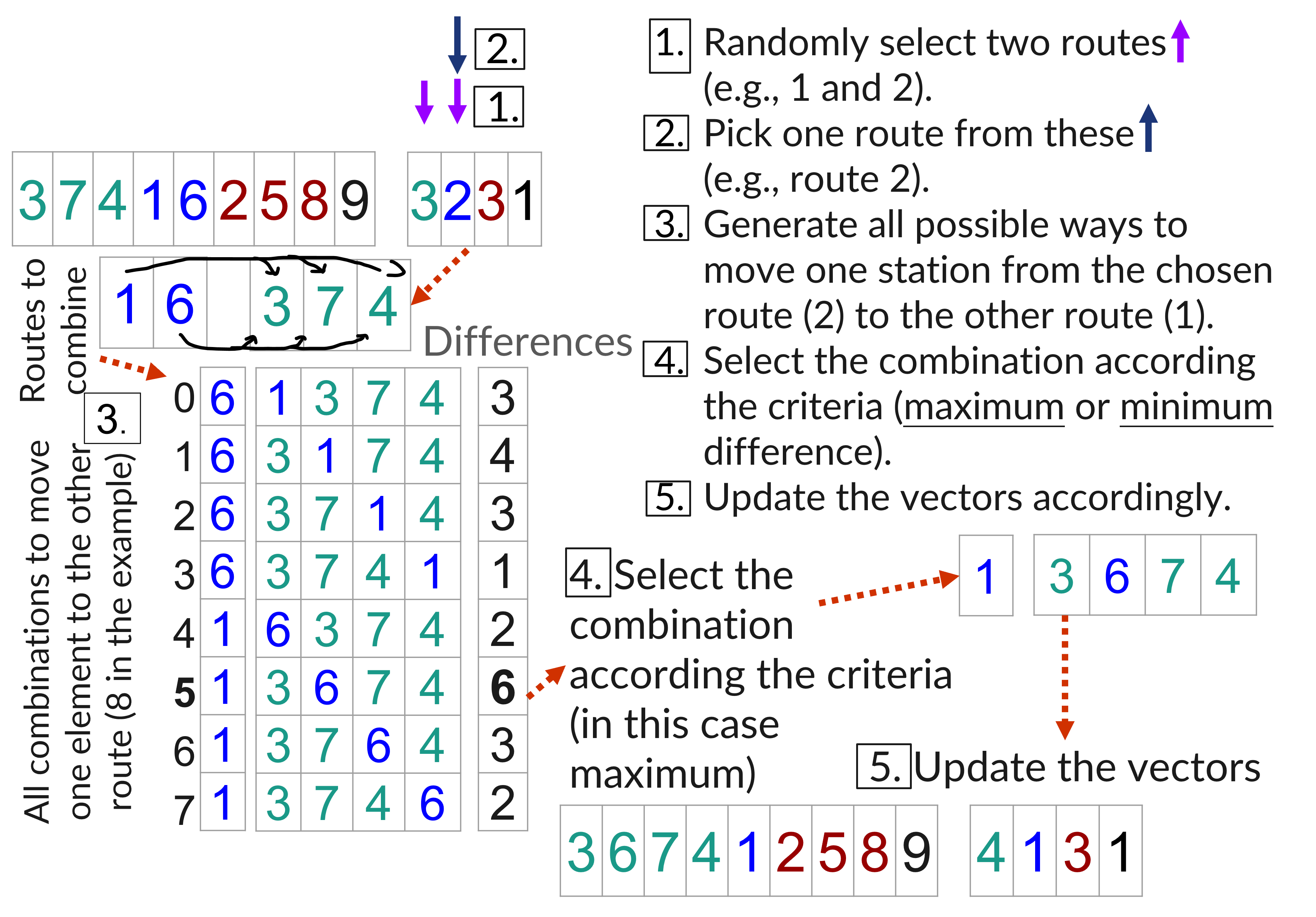}
\caption{BB1-MAX operator diagram. The red arrows indicate the step sequence.}
\label{mut:BB1}
\end{figure}

\paragraph{Best-improvement relocation.}
The \BBone (BB1-MIN and BB1-MAX) operator evaluates the 1--0 relocate neighborhood induced by a randomly selected pair of distinct routes. The operator sets one route as the source and the other as the destination, enumerates all feasible relocations of one station from the source to every insertion position in the destination, and selects the move that optimizes the induced change in travel distance in both trucks: BB1-MIN selects the minimum change, whereas BB1-MAX selects the maximum change. Distance evaluations use the route-distance function $\delta(\cdot)$ defined in Eq.~\eqref{eq:route_distance}. Algorithm~\ref{algo:BB1} and \figurename~\ref{mut:BB1} summarize the procedure; the diagram depicts BB1-MAX, and BB1-MIN differs only in using an argmin instead of an argmax.

\begin{algorithm}[!h]
\setlength{\abovecaptionskip}{0pt}
\setlength{\belowcaptionskip}{0pt}
\small
\caption{\BBone BB1-MIN / BB1-MAX)}\label{algo:BB1}
\begin{algorithmic}[1]
\State Randomly select two distinct routes.
\State Select one route as the source and set the other as the destination.
\State Generate all feasible single-station relocations from the source to the destination route.
\State Evaluate each relocation using the induced change in route distance $\delta(\cdot)$; select the best-improvement relocation (minimum for BB1-MIN, maximum for BB1-MAX).
\State Update the solution encoding (\routes,\ktrucks) according to the selected relocation.
\end{algorithmic}
\end{algorithm}

\paragraph{Biased relocation sampling.}
The \BBtwo (BB2-MIN and BB2-MAX) operator employs the same 1--0 relocate neighborhood as BB1 but selects relocations stochastically via roulette-wheel sampling biased toward low-distance-increase (BB2-MIN) or high-distance-increase (BB2-MAX) moves. Let $R_t$ and $R_{t'}$ denote the randomly selected routes, with $\mathcal{N}{rel}(R_t,R{t'})$ as the set of all feasible single-station relocations from $R_t$ to admissible insertion positions in $R_{t'}$. 

For each move $m\in\mathcal{N}{rel}(R_t,R{t'})$, the induced distance change $\Delta_m$ is computed as shown in Eq.~\eqref{eq:bb2_delta}.
The operator transforms $\{\Delta_m\}_{m\in\mathcal{N}_{rel}(R_t,R_{t'})}$ into roulette probabilities using a  $c_m$ parameter-free normalization: $c_m=\Delta_m-\min_{\ell\in\mathcal{N}_{rel}(R_t,R_{t'})}\Delta_{\ell}$ (in BB2-MIN) or $c_m=\Delta_m-\max_{\ell\in\mathcal{N}_{rel}(R_t,R_{t'})}\Delta_{\ell}$ (in BB2-MAX).

\begin{equation}
\Delta_m =
\bigl(\delta(R_t^{(m)})+\delta(R_{t'}^{(m)})\bigr)
-\bigl(\delta(R_t)+\delta(R_{t'})\bigr),
\label{eq:bb2_delta}
\end{equation}

\begin{figure}[!h]
\setlength{\abovecaptionskip}{0pt}
\setlength{\belowcaptionskip}{0pt}
\includegraphics[trim=0 0 0 2mm,clip,width=1\linewidth]{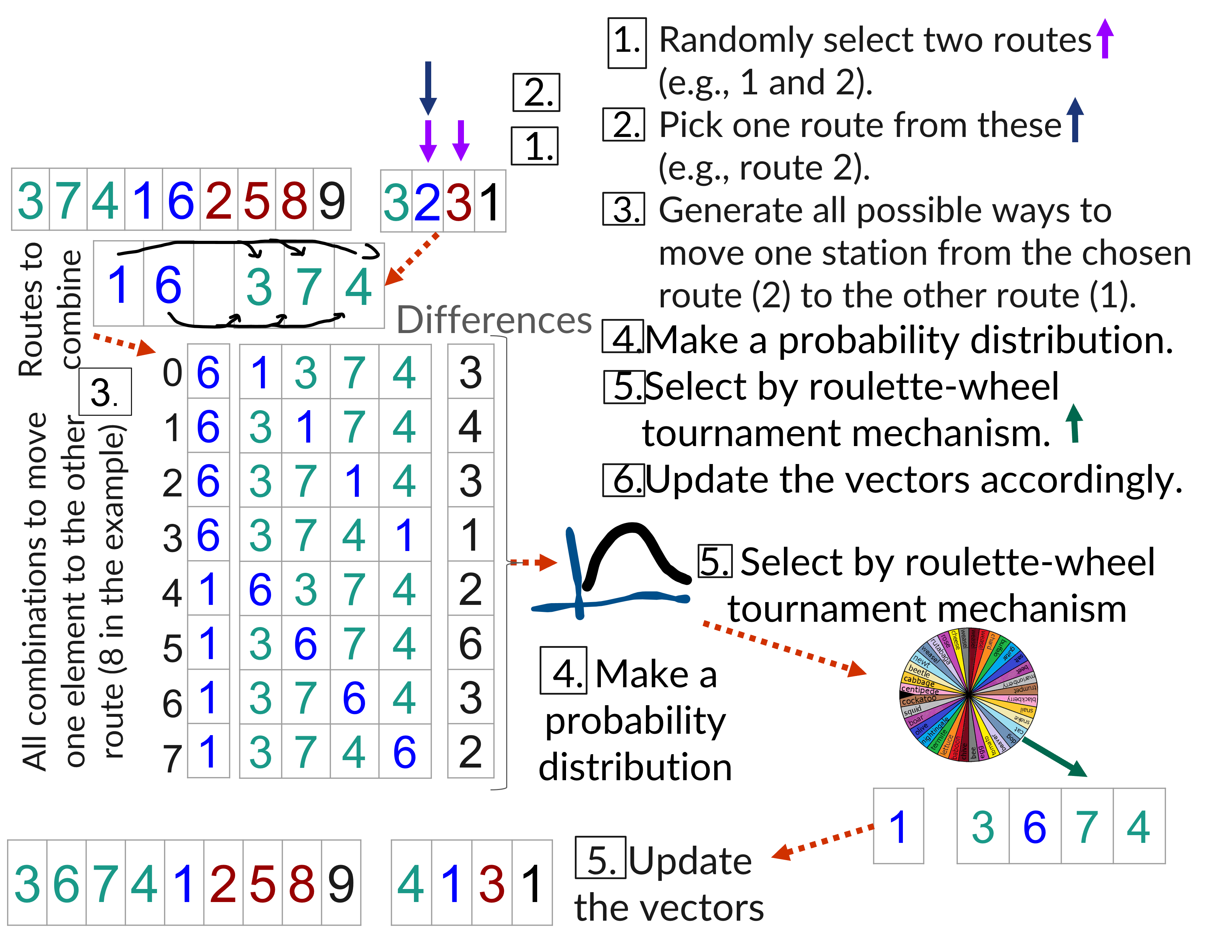}
\caption{Diagram of the \BBtwo{} operator (BB2-MIN/MAX).}
\label{mut:BB2}
\end{figure}

Weights ($w_m$) and probabilities ($P(m)$) are computed according to Eq.~\eqref{eq:bb2}.
The operator samples one move $m$ according to $P(m)$ and applies it to update (\routes,\ktrucks). Algorithm~\ref{algo:BB2} and \figurename~\ref{mut:BB2} summarize the procedure (red arrows indicate the step sequence).
\begin{equation}
\begin{split}
w_m&=\frac{1}{\varepsilon+c_m},\\
P(m)&=\frac{w_m}{\sum_{\ell\in\mathcal{N}{rel}(R_t,R{t'})} w_{\ell}},
\end{split}
\label{eq:bb2}
\end{equation}

\begin{algorithm}[!h]
\caption{\BBtwo (BB2-MIN / BB2-MAX)}\label{algo:BB2}
\begin{algorithmic}[1]
\State Randomly select two distinct routes.
\State Select one as source $R_t$ and the other as destination $R_{t'}$.
\State Generate $\mathcal{N}{rel}(R_t,R{t'})$: all feasible single-station relocations.
\State Compute ${\Delta_m}$ and roulette probabilities $P(m)$ per Eqs.~\eqref{eq:bb2_delta}--\eqref{eq:bb2} (min-shift for BB2-MIN, max-shift for BB2-MAX).
\State Sample relocation $m$ according to $P(m)$.
\State Update solution encoding (\routes,\ktrucks).
\end{algorithmic}
\end{algorithm}

The AB2 operator applies an unbiased relocation by selecting both the moved station and its insertion position uniformly at random.
The BB1-MIN/MAX operators apply a best-improvement relocation by choosing, among all candidate relocations between the selected routes, the move that minimizes/maximizes the induced change in travel distance.
The BB2-MIN/MAX operators use the same relocate neighborhood as BB1-MIN/MAX, but select a move by roulette-wheel sampling with probabilities derived from induced distance changes, which adds stochasticity while biasing selection toward relocations that preserve route length.

\section{Experimental Settings}
\label{sec:experimental-setup}

This section describes the case-study instance, the evaluation protocol, and the execution platform. The section also reports the parameter-tuning procedure used to set NSGA-II hyperparameters.

\subsection{Problem instance}
\begin{figure}[!ht]
\setlength{\abovecaptionskip}{0pt}
\setlength{\belowcaptionskip}{-6pt}
\centering
\includegraphics[width=.9\linewidth]{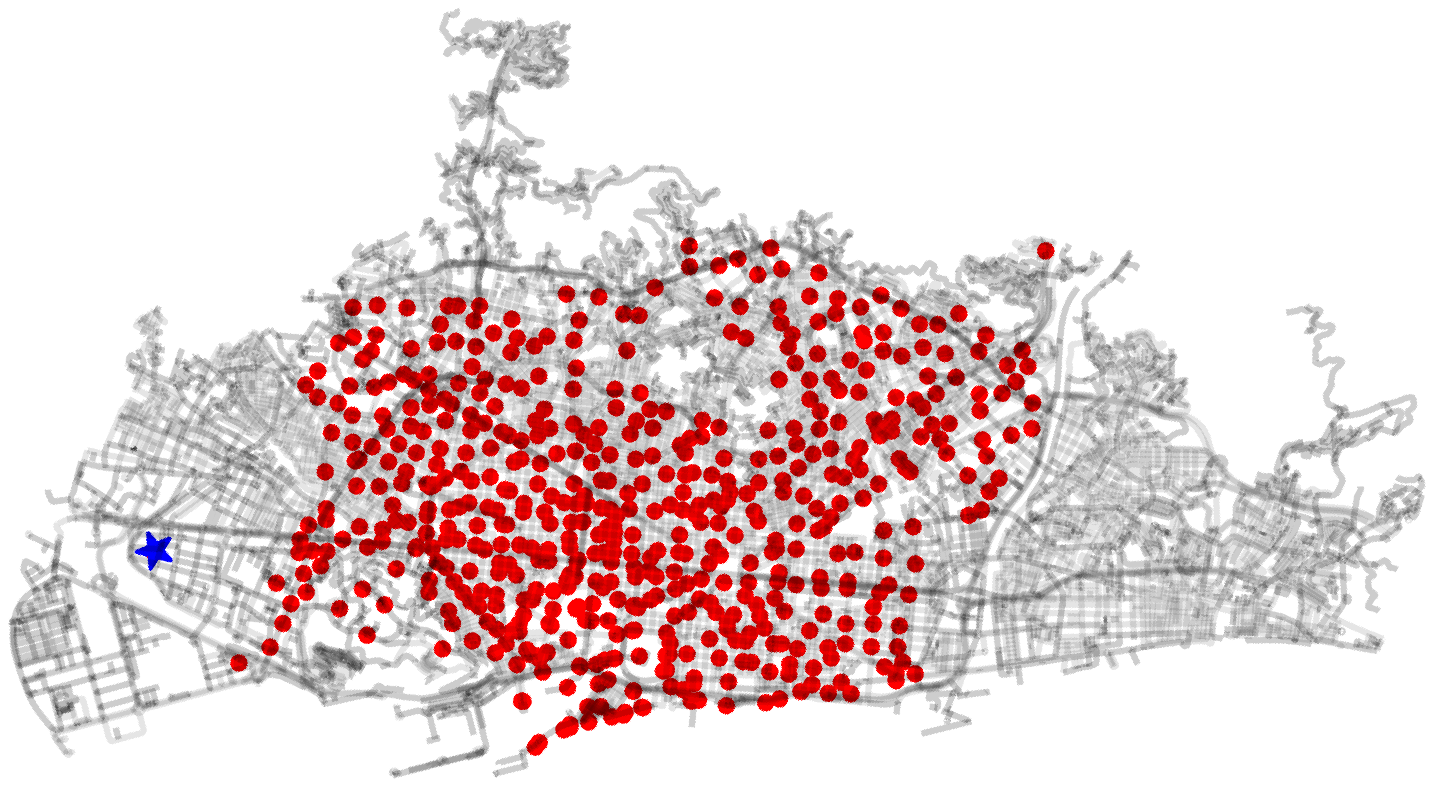}
\caption{Instance map: the red points are the bike-shared stations and the blue is the depot.} 
  \label{fig:mapBarceloan}
\end{figure}

Bike demand is inferred from changes in bicycle availability at each station and aggregated at an hourly resolution. The experiments focus on Mondays 07{:}00--08{:}00. The dataset covers 2019--2025, with station status sampled every 4--5 minutes. After preprocessing and missing-data handling, the instance includes 460 dock-based stations and 328 Mondays. Figure~\ref{fig:mapBarceloan} shows the station locations and the depot. The fleet contains 20~trucks with capacity 20~bikes each.

Demand uncertainty is represented with scenarios derived from historical observations. The data are split into 80\% training and 20\% validation. For each split, Monte Carlo sampling generates 15 training scenarios and 15 validation scenarios for each evaluation. The robustness objective uses a high-demand scenario, i.e., the 90th percentile of the empirical demand distribution for each station. Hyperparameter tuning is performed on a reduced instance containing 25\% of the stations.

\subsection{Evaluated metrics and baseline methods\label{sec:metrics}}

This subsection summarizes the evaluated metrics and heuristic greedy baselines.

\textit{Multiobjective optimization (MO) metrics.}
MO performance is measured with $gd+$ and $igd+$ for proximity to the Pareto front~\cite{AUDET2021397}. The $spread$ metric evaluates the dispersion of non-dominated solutions~\cite{spread}, and $\#nds$ counts the obtained non-dominated set size. The relative hypervolume ($rhv$) assesses coverage and dominance~\cite{hv}. A reference front is approximated by pooling all non-dominated solutions across runs because the true Pareto front is unknown.

\textit{Problem-related metrics.}
Operational performance is assessed via total travel distance and scenario-weighted unmet demand. Total distance is computed as $f_{\mathrm{dist}}(\mathbf{R})$ (Eq.~\eqref{eq:obj_distance}) from route distance $\delta(\cdot)$ (Eq.~\eqref{eq:route_distance}) using shortest-path distances from Dijkstra's algorithm. Scenario-weighted unmet demand is $f_{\mathrm{ud}}(\mathbf{R})$ (Eq.~\eqref{eq:obj_unmet}), where unmet demand is obtained from the deterministic recourse evaluation $\mathcal{Z}(\mathbf{R},h)$ (Section~\ref{sec:problem}); unvisited stations contribute $|\tau_{i,h}-O_i|$.

\textit{Greedy-based baseline methods.}
Two greedy constructive heuristics provide non-evolutionary baselines: RRCP-BI (Appendix~\ref{subsec:rapid}) and GLOBE (Appendix~\ref{subsec:globe}). Both construct feasible route sets and enforce single-visit constraints by design. Construction is guided by the expected-demand proxy $\widehat{D}_i$ (Eq.~\eqref{eq:rapid_proxy_need}), while objective values are computed by the deterministic evaluator. RRCP-BI builds pickup--drop pairs using a randomized restricted candidate list and assigns pairs to trucks via distance-aware balanced insertion. GLOBE samples feasible pair-moves from a global candidate set using score-biased probabilities. Baseline configurations target distance (GLOBE\_Dist, RRCP-BI\_Dist), demand satisfaction (GLOBE\_Serv, RRCP-BI\_Serv), or both (GLOBE\_SD, RRCP-BI\_SD).



\subsection{Parameter settings\label{subsec:parameter_settings}}
Parametric experiments identified the best-performing configurations of the domain-specific mutation operators (\ktrucks mutations) within NSGA-II. Tuning was performed with Irace~\cite{LOPEZIBANEZ201643}, which evaluates candidate configurations via iterative racing under a fixed budget. Relative hypervolume ($rhv$) was used as the tuning objective. Tuning used the 25\% instance with a budget of 1200 experiments.

Table~\ref{tab:parameters_hyper} lists the candidate parameter domains. Irace tunes the crossover and mutation operators applied to \routes and their probabilities. Population size is fixed to 200 and the number of generations to 500, based on preliminary experiments.



\begin{table}[!h]
\small
\renewcommand{\arraystretch}{0.9}
\caption{Candidates for the hyperparameters}
\label{tab:parameters_hyper}       
\begin{tabular}{ll}
\toprule
Parameter & Domain   \\
\midrule
Crossover on \routes ($C_R$) & OX, PMX, ERX, EERX \\
Mutation on \routes  ($M_R$)& IM, BMM, BSM, SM  \\
Probability of mutation \routes  ($PM_R$) & \{0.01, 0.05, 0.1, 0.15, 0.20, 0.25\} \\ 
Probability of crossover \routes  ($PC_R$) & \{0.1, 0.25, 0.5, 0.7, 0.9, 1\}  \\
Probability of mutation  \ktrucks  ($PM_T$) & \{0.01, 0.05, 0.1, 0.15, 0.20, 0.25\}  \\
\bottomrule
\end{tabular}
\end{table}

Table~\ref{tab:parameter_configs_nsga2} reports the best configuration selected for each mutation operator. Each selected configuration is evaluated with 36 independent runs on the 100\% instance for subsequent analysis.

\begin{table}[h!]
\small
\centering
\renewcommand{\arraystretch}{0.9}
\caption{Parameter configurations for each operator.}
\label{tab:parameter_configs_nsga2}
\begin{tabular}{lllrrr}
\toprule
Operator  & $C_R$ & $M_R$ & $PM_R$ & $PC_R$ & $PM_T$ \\
\midrule

AB2  & PMX & SM & 0.25 & 0.1 & 0.15  \\ 

BB1-MIN  & PMX & SM & 0.25 & 0.1 & 0.20  \\  

BB1-MAX  & PMX & BSM & 0.01 & 0.1 & 0.25  \\ 

BB2-MAX  & PMX & SM & 0.20 & 0.1 & 0.25  \\ 
BB2-MIN  & PMX & SM & 0.20 & 0.1 & 0.15  \\ 

\bottomrule
\end{tabular}
\end{table}

\subsection{Execution platform and implementation }
The proposed methods were developed in Python 3.10.12, using the scikit-learn, NetworkX, OSMnx~\citep{Boeing_2017_OSMNX}, and PyMOO~\citep{PyMOOarticle} libraries. Computational experiments were executed on a high-performance cluster consisting of 126 SD530 nodes, each equipped with 52 Intel Xeon Gold 6230R cores operating at 2.10 GHz and 192 GB of RAM. The nodes are connected through an InfiniBand HDR100 network, and each node includes 950 GB of local scratch storage. The source code is available at \url{https://github.com/pedrozad/bike-gecco-2026}. 


\section{Experimental analysis}
\label{sec:experimental-analysis}

This section presents the experimental results obtained with the proposed approach. All comparative results have been statistically validated using the Wilcoxon signed-rank test with Bonferroni correction and a significance level of $\alpha$ = 0.01; detailed test outcomes are reported in Appendix~\ref{app:statistics}. The analysis is organized into three parts: an evaluation of the proposed mutation operators, a validation against baseline approaches, and a problem-centric analysis assessing the practical impact of the obtained solutions.

\subsection{Analysis of the Proposed Operators}\label{subsec:analysis-proposal}

\begin{figure}[!h]
\setlength{\abovecaptionskip}{0pt}
\setlength{\belowcaptionskip}{0pt}
    \centering
    \includegraphics[trim=5mm 20mm 0mm 5mm, clip,width=0.75\linewidth]{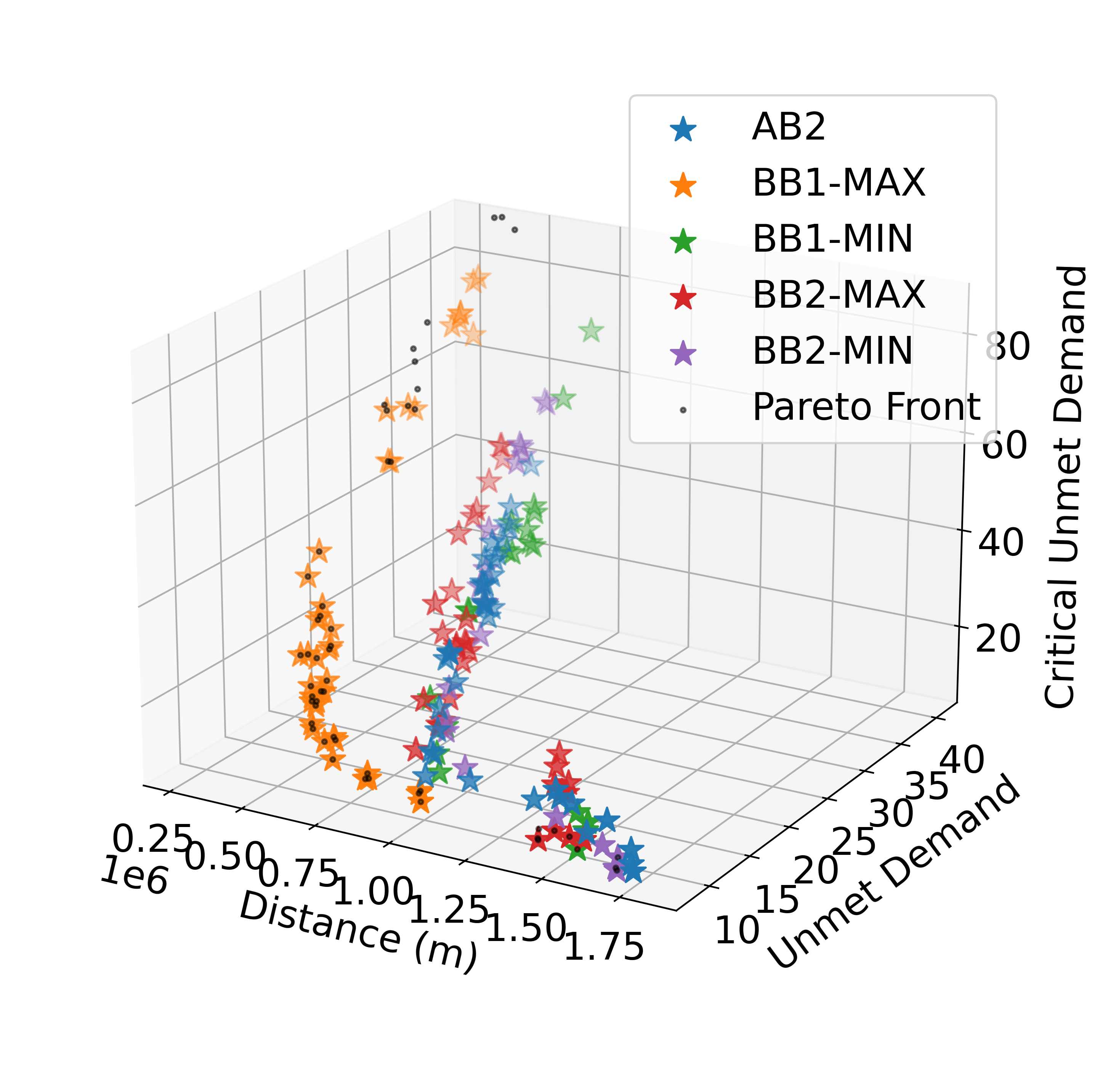}
    \caption{Three-dimensional Pareto front obtained by the different mutation operators.}
    \label{fig:pf-3d}
\end{figure}

\begin{figure}[!h]
\setlength{\abovecaptionskip}{0pt}
\setlength{\belowcaptionskip}{0pt}
    \centering
    \includegraphics[trim=0 2mm 0 0mm, clip,width=0.8\linewidth]{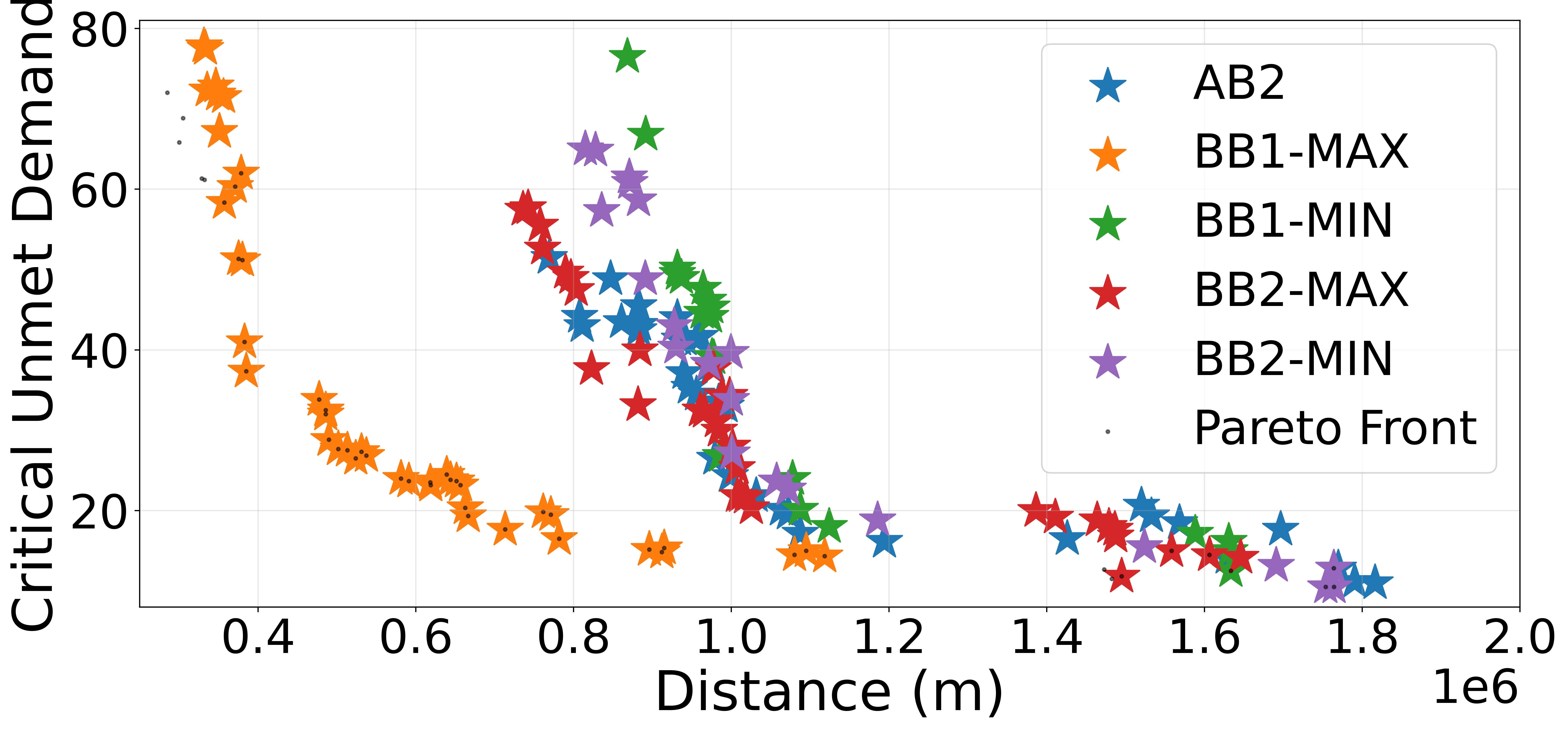}
    \caption{2D front projection: distance vs. critical unmet demand.}
    \label{fig:pf-f1-f3}
\end{figure}

This section analyzes the performance of the proposed mutation operators using both quantitative metrics and visual inspection of the resulting Pareto fronts. Tables \ref{tab:rhv-metrics}-\ref{tab:spread-metrics-nds} summarize the distribution of quality indicators across independent runs, while Figures~\ref{fig:pf-3d}–\ref{fig:pf-f1-f3} illustrate the corresponding Pareto fronts in the objective space, allowing a qualitative assessment of convergence and diversity.


In Table~\ref{tab:rhv-metrics}, the relative hypervolume (rhv, maximization) is reported. This metric evaluates the overall quality of the approximation set by jointly capturing convergence, dominance, and diversity with respect to a reference Pareto front, constructed by aggregating the non-dominated solutions obtained across all the experiments.

\begin{table}[!h]
\small
\renewcommand{\arraystretch}{0.9}
\caption{Results of rhv for the studied instance.}
\label{tab:rhv-metrics}
\begin{tabular}{lrrrr}
\toprule
 & Min & Median & IQR & Max \\
\midrule
AB2 & 0.445 & 0.504 & 0.035 & 0.533 \\
BB1-MAX & 0.620 & \textbf{0.750} & 0.054 & 0.851 \\
BB1-MIN & 0.360 & 0.438 & 0.044 & 0.530 \\
BB2-MAX & 0.388 & 0.503 & 0.040 & 0.551 \\
BB2-MIN & 0.382 & 0.467 & 0.039 & 0.541 \\
\bottomrule
\end{tabular}
\end{table}

The results show that BB1-MAX clearly outperforms all other variants, achieving the highest median rhv value as well as the best maximum performance. This indicates that promoting diversification through high-impact relocations, while retaining a deterministic best-improvement selection, is particularly effective for exploring high-quality trade-offs in the objective space.

In contrast, intensification-oriented operators (BB1-MIN and BB2-MIN) obtain substantially lower rhv values, suggesting that favoring low-impact moves limits exploration and restricts access to well-dominated regions of the Pareto front. Purely exploratory or stochastic strategies (AB2 and BB2 variants) achieve intermediate rhv values, confirming that exploration is beneficial but insufficient on its own to match the performance of a guided diversification strategy. No clear advantage is observed between purely random exploration (AB2) and biased stochastic sampling (BB2) in terms of rhv. The notable exception is BB1-MAX, where deterministic selection combined with explicit diversification yields a substantial performance gain over all stochastic alternatives. This highlights that, for this problem, how diversification is guided is more relevant than the degree of randomness itself.

Table~\ref{tab:igd-gd-metrics} reports the results for igd+ and gd+ (both minimization), which assess the convergence with respect to the reference Pareto front. While igd+ emphasizes coverage of the front, gd+ focuses on the average proximity of the obtained solutions.

\begin{table}[!h]
\setlength{\tabcolsep}{3pt}
\renewcommand{\arraystretch}{0.9}
\centering
\small
\caption{Results for IGD+ and GD+ metrics.}
\label{tab:igd-gd-metrics}
\begin{tabular}{lcccccccc}
\toprule
 & \multicolumn{4}{c}{IGD+} & \multicolumn{4}{c}{GD+} \\
\cmidrule(r){2-5} \cmidrule(l){6-9}
 & Min & Med & IQR & Max & Min & Med & IQR & Max \\
\midrule
AB2     & 0.156 & 0.181 & 0.014 & 0.210 & 0.110 & 0.141 & 0.029 & 0.186 \\
BB1-MAX & 0.043 & \textbf{0.065} & 0.009 & 0.093 & 0.062 & \textbf{0.078} & 0.006 & 0.102 \\
BB1-MIN & 0.179 & 0.217 & 0.027 & 0.252 & 0.114 & 0.159 & 0.025 & 0.200 \\
BB2-MAX & 0.144 & 0.176 & 0.016 & 0.208 & 0.116 & 0.151 & 0.027 & 0.183 \\
BB2-MIN & 0.168 & 0.203 & 0.023 & 0.249 & 0.106 & 0.152 & 0.029 & 0.199 \\
\bottomrule
\end{tabular}
\end{table}

The results are consistent across both metrics and strongly align with the rhv analysis. BB1-MAX achieves the lowest median values for both IGD+ and GD+, indicating superior convergence and a closer approximation to the reference front. This confirms that deterministic best-improvement combined with diversification not only improves dominance but also enhances convergence quality. Exploratory operators such as AB2 and BB2-MAX systematically outperform intensification-oriented variants, reinforcing the observation that exploration is essential to avoid premature convergence. In contrast, BB1-MIN exhibits the worst convergence behavior, despite producing compact solution sets, suggesting that excessive intensification restricts the search to suboptimal solution regions.

Table~\ref{tab:spread-metrics-nds} reports the results for spread and \#nds, two indicators that characterize the structure of the obtained Pareto fronts but do not directly assess their quality with respect to the reference front. The spread metric measures the dispersion of solutions around their centroid and is therefore a front-internal indicator, independent of convergence or dominance. Under this definition, BB1-MIN achieves the highest median spread, indicating more dispersed solution sets. However, this dispersion does not translate into better performance, as these solutions remain far from the reference Pareto front, a fact already evidenced by the IGD+ and GD+ results. In contrast, BB1-MAX exhibits lower spread values, reflecting more compact solution sets that are nonetheless better aligned with the reference front. This indicates that high-quality fronts do not require large internal dispersion, but rather an effective placement of solutions along the true trade-off surface.

\begin{table}[!h]
\setlength{\tabcolsep}{4pt}
\renewcommand{\arraystretch}{0.9}
\centering
\small
\caption{Results for Spread and \#NDS metrics.}
\label{tab:spread-metrics-nds}
\begin{tabular}{lcccccccc}
\toprule
 & \multicolumn{4}{c}{Spread} & \multicolumn{4}{c}{\#NDS} \\
\cmidrule(r){2-5} \cmidrule(l){6-9}
 & Min & Med & IQR & Max & Min & Med & IQR & Max \\
\midrule
AB2     & 5.074 & 5.680 & 0.481 & 6.142 & 194 & 198 & 2.25 & 200 \\
BB1-MAX & 2.247 & 2.576 & 0.456 & 4.126 & 188 & 199 & 2.00 & 200 \\
BB1-MIN & 5.583 & \textbf{6.462} & 0.390 & 6.991 & 192 & 198 & 2.00 & 200 \\
BB2-MAX & 4.911 & 5.348 & 0.346 & 5.904 & 190 & 198 & 1.25 & 200 \\
BB2-MIN & 5.121 & 5.981 & 0.480 & 6.642 & 190 & 198 & 2.00 & 200 \\
\bottomrule
\end{tabular}
\end{table}

Regarding the number of non-dominated solutions (\#nds), all operators produce similarly large non-dominated sets, with median values close to the population size. Statistical analysis (Appendix~\ref{app:statistics}) indicates that no statistically significant differences are observed for this metric. This suggests that all variants are equally capable of maintaining large non-dominated populations, and that performance differences are primarily driven by solution quality and front location, rather than by the cardinality of the obtained sets.


To complement the quantitative analysis, we now provide a visual validation of the proposed operators by inspecting the resulting Pareto fronts in the objective space. We include a three-dimensional representation of the complete front to assess global convergence and coverage (Figure~\ref{fig:pf-3d}), together with a two-dimensional projection highlighting the most relevant trade-off of the problem under critical demand conditions (Figure~\ref{fig:pf-f1-f3}). Similar trends are observed when considering expected unmet demand, and therefore only the critical case is reported.

Figure~\ref{fig:pf-3d} shows the three-dimensional Pareto fronts obtained by the different mutation operators together with the reference front. The visualization confirms the quantitative results: BB1-MAX consistently produces solution sets that are closer to the reference front across the three objectives, whereas intensification-oriented variants converge to regions that are visibly shifted away from the true trade-off surface. Exploratory strategies achieve broader coverage but exhibit weaker proximity to the reference front.



Figure~\ref{fig:pf-f1-f3} illustrates the trade-off between travel distance and critical unmet demand, directly reflecting the robustness-oriented objective introduced in this work. The projection confirms the superior alignment of BB1-MAX with the reference Pareto front, while intensification-oriented variants remain shifted away from the true trade-off surface. Other operators are able to reach only partial segments of the front, typically corresponding to solutions that require large increases in distance for marginal improvements in service quality. These observations are consistent across both service quality objectives, reinforcing the conclusion that guided diversification enables the identification of well-balanced and meaningful solutions.

\subsection{Validation against Baseline Approaches}\label{subsec:validation}

This section validates our approach by comparing BB1-MAX against several greedy baselines and an ablated variant where the domain-specific mutation is replaced by a standard inversion operator.

Table~\ref{tab:mo-algorithms-metrics} summarizes the results. Here, \#nds denotes the number of non-dominated solutions contributed by each method to the global reference Pareto front. This metric reflects the effective contribution of each approach to the overall trade-off surface.

\begin{table}[!h]
\caption{Results of MO metrics.}
\label{tab:mo-algorithms-metrics}
\centering
\small
\begin{tabular}{lrrrrr}
\toprule
 & rhv & gd+ & igd+ & spread & \#nds \\
\midrule
BB1-MAX & \textbf{0.975} & 0.004 & \textbf{0.007} & 1.241 & \textbf{36} \\
BB1-MIN & 0.681 & 0.092 & 0.153 & 1.564 & 1 \\
BB2-MAX & 0.742 & 0.062 & 0.111 & \textbf{2.046} & 3 \\
BB2-MIN & 0.696 & 0.087 & 0.142 & 1.699 & 3 \\
\hline
GLOBE\_SD & 0.402 & 0.007 & 0.127 & 0.414 & 2 \\
GLOBE\_Dist & 0.251 & 0.048 & 0.180 & 0.377 & 0 \\
GLOBE\_Serv & 0.362 & \textbf{0} & 0.131 & 0.294 & 6 \\
RRCP-BI\_SD & 0.283 & 0.096 & 0.172 & 0.378 & 0 \\
RRCP-BI\_Dist & 0.279 & 0.098 & 0.181 & 0.423 & 0 \\
RRCP-BI\_Serv & 0.294 & 0.086 & 0.171 & 0.442 & 0 \\
\hline
BB1-MAX\_abl & 0.674 & 0.070 & 0.152 & 2.420 & 2 \\
\bottomrule
\end{tabular}
\end{table}

The results show that BB1-MAX clearly dominates the comparison, achieving the best convergence and dominance metrics while contributing a substantially larger portion of the reference Pareto front. In contrast, greedy strategies are only able to contribute solutions located at extreme regions of the front, typically characterized by very low distance but extremely poor service quality (high unmet and critical unmet demand).


The ablated evolutionary variant exhibits a behavior similar to the remaining evolutionary baselines: it is able to identify a small number of isolated non-dominated solutions, but these correspond to large increases in distance with only marginal improvements in service quality, resulting in a limited contribution to the front.

\begin{figure}[!t]
\setlength{\abovecaptionskip}{0pt}
\setlength{\belowcaptionskip}{0pt}
    \centering
    \includegraphics[width=0.75\linewidth]{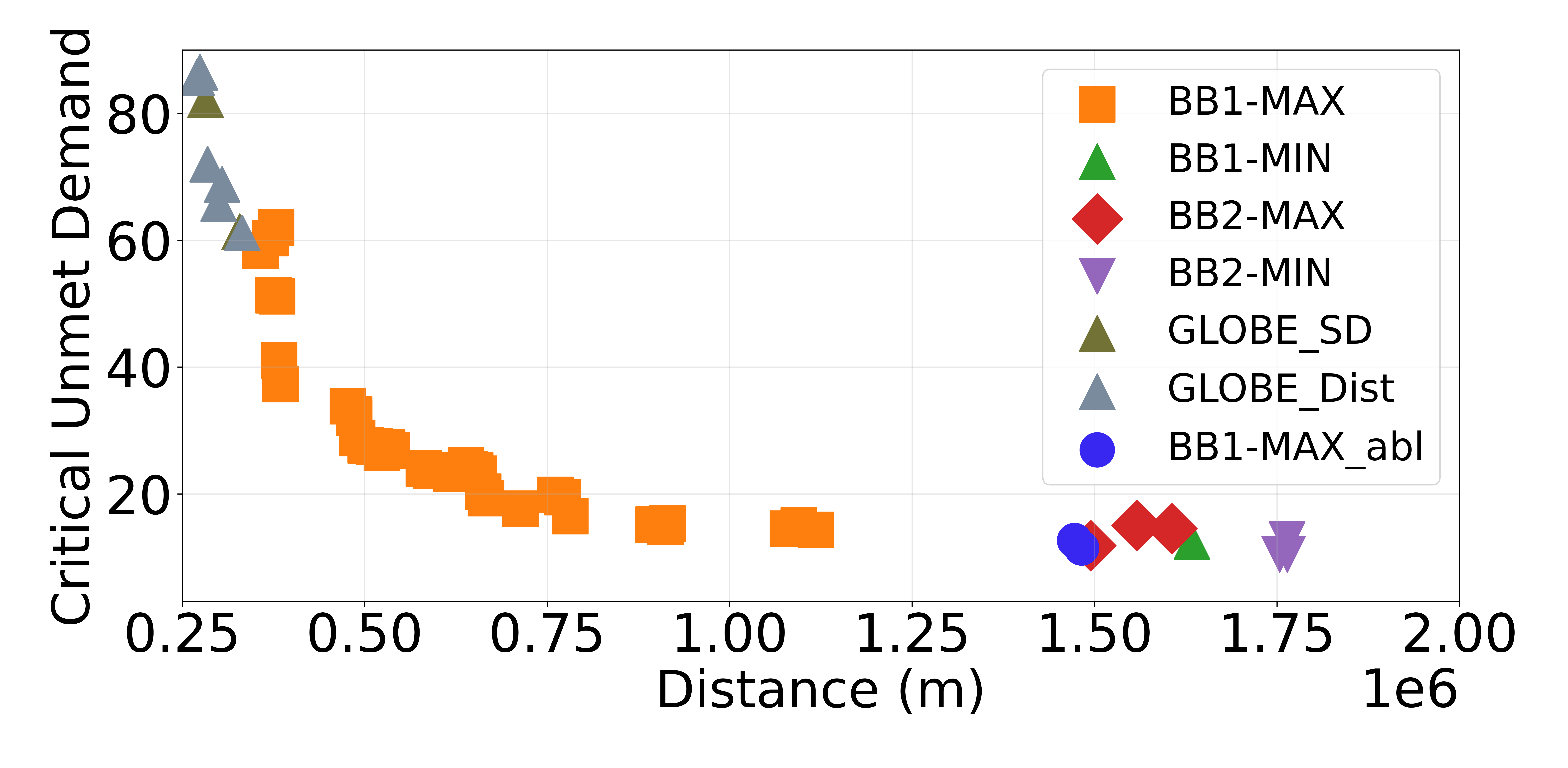}
    \caption{Two-dimensional Reference Pareto front projection: distance vs. critical unmet demand.}
    \label{fig:pf2-f1-f3}
\end{figure}

Figure~\ref{fig:pf2-f1-f3} visually confirms these observations by showing the reference Pareto front together with the origin of each solution. Notably, only two greedy methods contribute non-dominated solutions, both sharing a strong bias toward distance minimization, while the remaining greedy approaches fail to reach competitive trade-offs. Overall, these results highlight that the advantage of BB1-MAX lies not only in reaching the Pareto front, but in identifying well-balanced and practically relevant regions of it, rather than extreme and poorly compensated solutions.

\subsection{Problem-centric analysis}\label{subsec:domain-analysis}


Finally, this section analyzes the obtained solutions from a problem-centric perspective, focusing on their operational impact and comparing them with the current system (no rebalancing). We first assess service availability at station level, and then study the trade-offs induced by different fleet sizes in the solutions generated.

\begin{figure}[!h]
\setlength{\abovecaptionskip}{0pt}
\setlength{\belowcaptionskip}{0pt}
\includegraphics[trim=0 0 0 0mm, clip,width=0.75\linewidth]{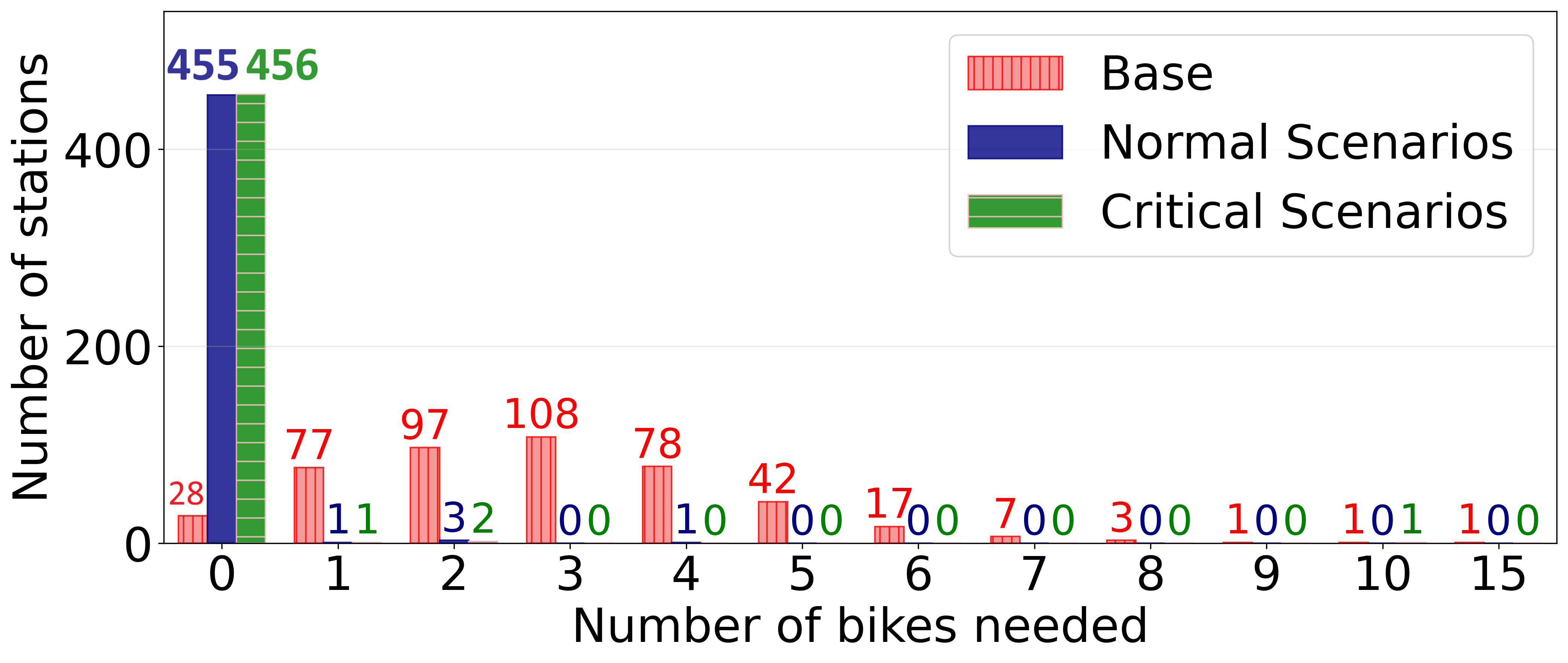}
\caption{Histogram of the number of bikes needed per station in the non-rebalancing configuration (BASE), compared with the closest to ideal vector in normal and critical cases.}
\label{hist:robust}
\end{figure}

Figure~\ref{hist:robust} reports the distribution of stations according to the number of bikes that users attempted to obtain but were unavailable. Results are shown for the current configuration and for the BB1-MAX solution closest to the ideal point, evaluated under both normal and critical demand scenarios. In the non-rebalancing configuration, only a limited number of stations fully satisfy demand, while a large fraction experience shortages. In contrast, the proposed solution substantially increases the number of stations with zero unmet demand, indicating a clear improvement in service quality. In fact, this behavior is preserved under critical cases, showing that the selected solution is robust to high-demand conditions.

\begin{figure}[!h]
\setlength{\abovecaptionskip}{0pt}
\setlength{\belowcaptionskip}{0pt}
\includegraphics[trim=0 0 0 0mm, clip, width=0.75\linewidth]{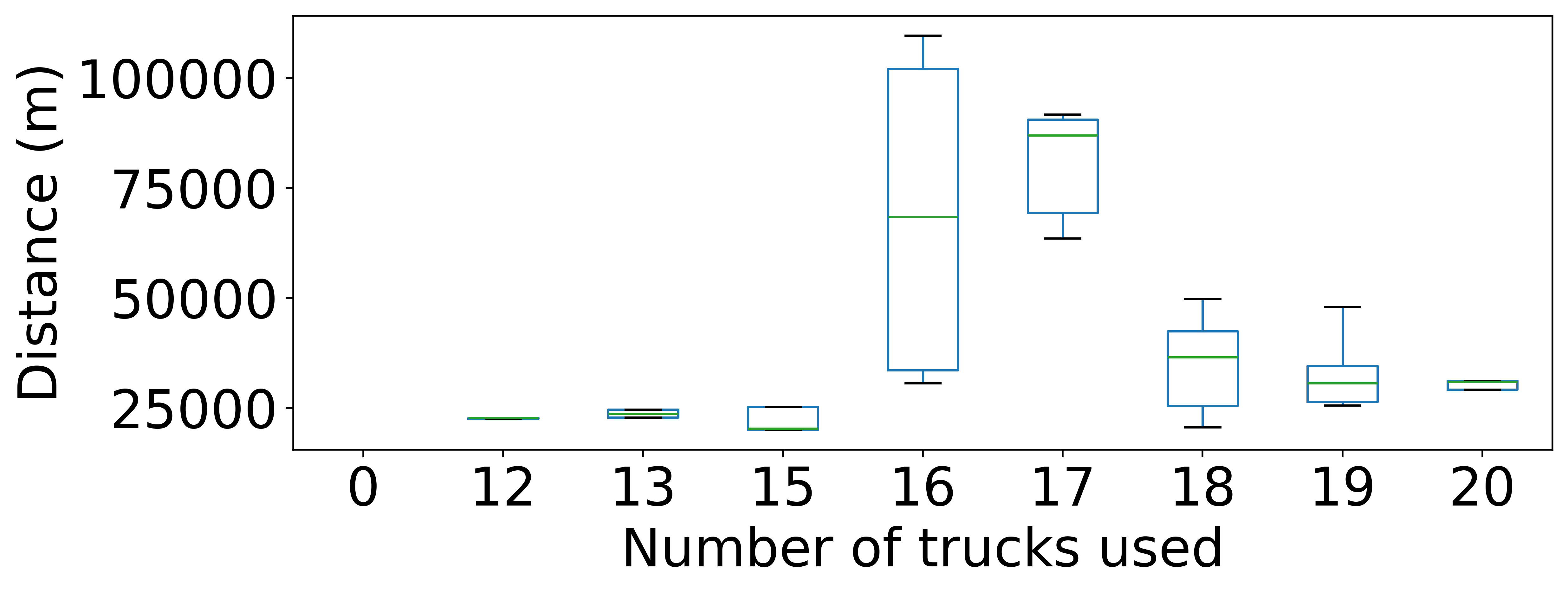}
\caption{Total distance distribution of the solutions of our approach by number of trucks.}
\label{fig:truck_vs_distance}
\end{figure}

\begin{figure}[!h]
\includegraphics[width=0.75\linewidth]{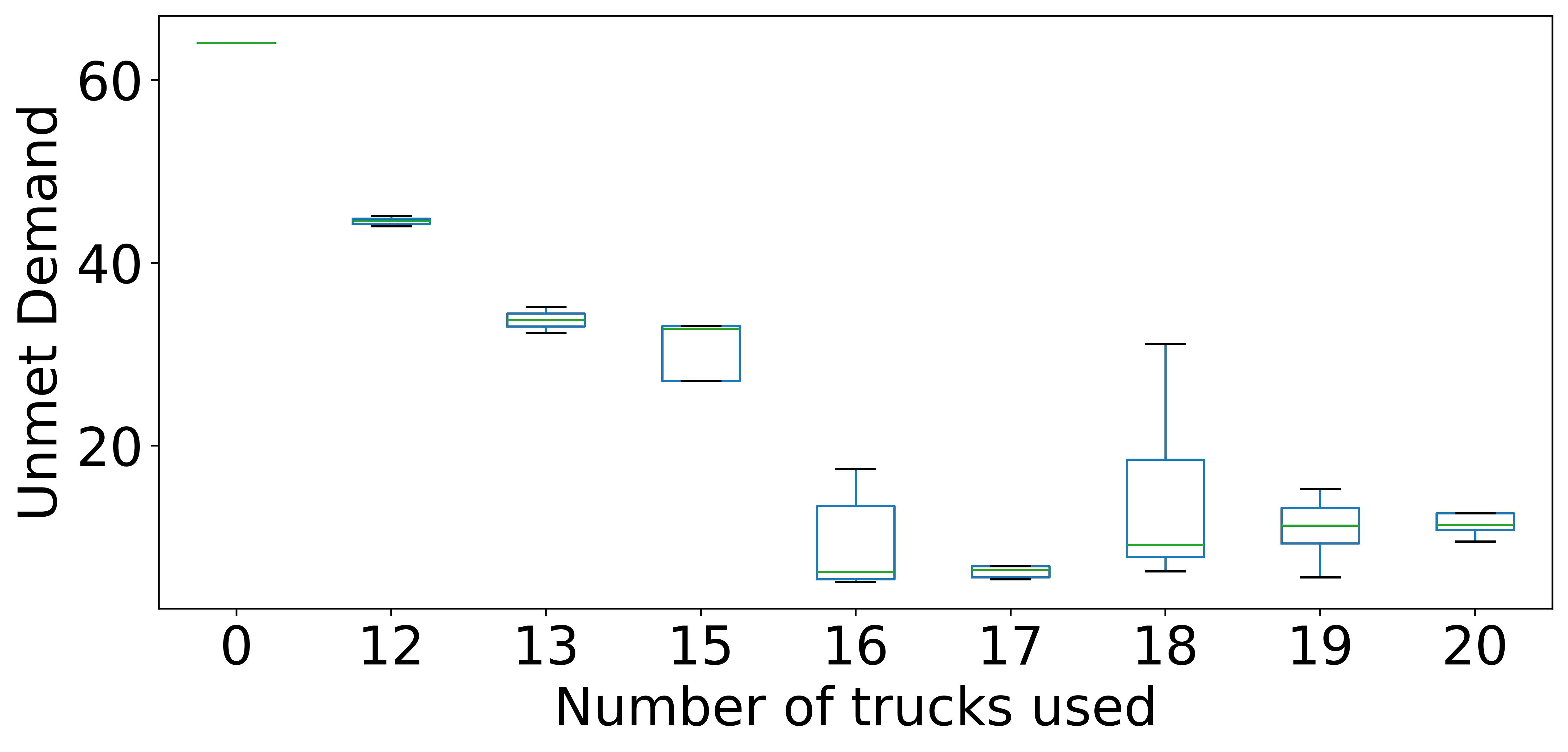}
\caption{Unmet demand distribution of the solutions of our approach by number of trucks.}
\label{fig:truck_vs_demmand}
\end{figure}

Figures~\ref{fig:truck_vs_distance} and~\ref{fig:truck_vs_demmand} analyze the Pareto-optimal solutions produced by BB1-MAX by grouping them according to the number of trucks used. For each fleet size, boxplots summarize the distribution of total distance and unmet demand (the critical unmet demand is very similar). The case with zero trucks corresponds to the current system configuration, which exhibits zero distance but the highest unmet demand. Solutions using up to 15 trucks achieve very low travel distances, but at the cost of poor service quality. The transition to 16 trucks represents a clear improvement in unmet demand, although with higher variability in distance. Solutions with 17 trucks further stabilize distance while maintaining good service levels, whereas additional trucks beyond this point yield only marginal improvements in service quality and distance variability. These results suggest that 16–18 trucks provide the best compromise between operational cost and service quality, with diminishing returns observed for larger fleet sizes.

\section{Conclusions and future work}
\label{sec:conclusions}

This work proposed a set of domain-specific mutation operators for a tri-objective bike rebalancing problem, together with a novel formulation of a robustness-oriented objective that explicitly accounts for service degradation under critical demand scenarios. By integrating this third objective, the approach moves beyond average-case optimization and enables the systematic identification of solutions that remain effective under adverse conditions.

The experimental analysis shows that guided diversification, particularly through the BB1-MAX operator, significantly improves convergence, dominance, and practical solution quality when compared to greedy strategies and ablated variants. The obtained solutions not only perform well in multi-objective metrics, but also exhibit robust and operationally meaningful trade-offs, improving service availability while keeping distance at reasonable levels.

Future work will focus on three main directions. First, the proposed operators will be assessed within alternative MOEAs to evaluate their generality. Second, additional adaptive and hybrid domain-aware operators will be explored to enhance search efficiency. Finally, the modeling of demand will be extended using machine learning techniques to capture temporal dynamics and uncertainty.

\begin{acks}
This research is partially funded by the grant number PID2024-158752OB-I00 (AIM-ZERO) by MICIU/AEI/10.13039/501100011033; and the grant number PRE2021-100645 by MCIN/AEI/10.13039/501100011033 and by the FSE+. The authors thank the Supercomputing and Bioinformatics center (UMA) for their computer resources and assistance. 
\end{acks}

\bibliographystyle{plain}
\bibliography{bibliography}

\clearpage
\appendix

\section{Greedy constructive baselines}\label{sec:greedy_baselines}

This section describes two greedy constructive heuristics devised as baselines for the tri-objective overnight rebalancing problem in Section~\ref{sec:problem}.
Both heuristics output a feasible route set $\mathbf{R}=\{R_t\}_{t\in\mathcal{T}}$ (Eq.~\eqref{eq:route_def}) and satisfy the single-visit constraints by construction (Eqs.~\eqref{eq:unique_service}--\eqref{eq:no_repeat_within_route}).
Both heuristics use an \emph{expected-demand proxy} to guide construction, but they are not scenario-aware optimizers: the objective values are computed exclusively by the deterministic scenario evaluator in Section~\ref{sec:evaluation}.

\subsection{Randomized Restricted Candidate Pairing with Balanced Insertion (RRCP-BI)}\label{subsec:rapid}

RRCP-BI is a two-stage greedy constructor designed to address the problem for large instances.
Stage~1 stochastically creates pickup--drop pairs under a restricted candidate list (RCL) policy.
Stage~2 assigns pairs to trucks using a distance-aware insertion rule to obtain $\mathbf{R}$.

RRCP-BI derives a deterministic proxy net desired change from an expected-demand input $\widehat{D}_i$ and the corresponding proxy deficit $\mathrm{def}_i$ and surplus $\mathrm{sur}_i$ magnitudes according to Eq.~\eqref{eq:rapid_proxy_need} and Eq.~\eqref{eq:rapid_surdef}, respectively.
\begin{equation}
\widehat{D}_i \;=\; \texttt{expected\_demand}_i - O_i,
\label{eq:rapid_proxy_need}
\end{equation}
\begin{align}
\mathrm{def}_i \;=\; \max\{\widehat{D}_i,0\},\qquad
\mathrm{sur}_i \;=\; \max\{-\widehat{D}_i,0\}.
\label{eq:rapid_surdef}
\end{align}

Let $\mathcal{S}^{+}$ and $\mathcal{S}^{-}$ be the sets of deficit and surplus stations, respectively (Eq.~\eqref{eq:def_sur}).
For a candidate pair $(i,j)$ with $i\in\mathcal{S}^{-}$ and $j\in\mathcal{S}^{+}$, RRCP-BI defines a capacity-capped proxy transfer volume $q(i,j)$ according to Eq.~\eqref{eq:rapid_q}, where $C$ is the truck capacity (in bikes). This value is used only for construction; scenario-dependent transfers and unmet demand are computed by the evaluator (Section~\ref{sec:evaluation}).

\begin{align}
\mathcal{S}^{+}=\{i\in\mathcal{S}\mid \mathrm{def}_i>0\},\qquad
\mathcal{S}^{-}=\{i\in\mathcal{S}\mid \mathrm{sur}_i>0\}.
\label{eq:def_sur}
\end{align}

\begin{equation}
q(i,j)\;=\;\min\{\mathrm{sur}_i,\mathrm{def}_j,C\}.
\label{eq:rapid_q}
\end{equation}

Algorithm~\ref{alg:rapid} summarizes RRCP-BI. The method consists of Stage~1 (randomized pairing, lines~\ref{ln:rrcp:while}--\ref{ln:endwhile1}) and Stage~2 (pair insertion into truck routes, lines~\ref{ln:rrcp:init_routes}--\ref{ln:endfor2}).

\paragraph{Parameters}
RRCP-BI uses six user parameters.
$m_{\max}$ is the maximum size of the restricted candidate list $\mathrm{RCL}(i)$ and bounds the number of deficit candidates evaluated per iteration.
$\beta_{\mathrm{pick}}\ge 0$ controls pickup selection pressure in Eq.~\eqref{eq:rapid_pick_prob}: $\beta_{\mathrm{pick}}=0$ yields uniform sampling over $\mathcal{A}^{-}$, and larger values bias selection toward larger surplus $\mathrm{sur}_i$.
$\beta_{\mathrm{drop}}\ge 0$ controls drop selection pressure in Eq.~\eqref{eq:rapid_drop_prob}: $\beta_{\mathrm{drop}}=0$ yields uniform sampling within $\mathrm{RCL}(i)$, and larger values bias selection toward higher $\mathrm{score}(i,j)$.
The \emph{score mode} selects the definition of $\mathrm{score}(i,j)$ (service-only, inverse-distance, or service-per-distance; Eqs.~\eqref{eq:rapid_score_svc}--\eqref{eq:rapid_score_spd}), which determines the construction bias toward transfer volume versus proximity.
$\lambda\ge 0$ is the load-balancing penalty in Eq.~\eqref{eq:rapid_balanced_cost} that discourages assigning many pairs to the same truck by increasing the cost of trucks with larger $n_t$.
$\varepsilon>0$ is a numerical-stability constant used in Eqs.~\eqref{eq:rapid_pick_prob} and \eqref{eq:rapid_drop_prob} and in the inverse-distance terms to avoid division by zero and undefined probabilities.

\paragraph{Stage 1: randomized pair-and-remove with RCL} 
RRCP-BI computes the proxy quantities $\widehat{D}_i$, $\mathrm{sur}_i$, and $\mathrm{def}_i$ (Alg.~\ref{alg:rapid}, line~\ref{ln:rrcp:proxy}) and initializes the available sets $\mathcal{A}^{-}$ and $\mathcal{A}^{+}$ and the pair list $\mathcal{P}$ (Alg.~\ref{alg:rapid}, line~\ref{ln:rrcp:init_sets}).
RRCP-BI iterates while $\mathcal{A}^{-}\neq\emptyset$ and $\mathcal{A}^{+}\neq\emptyset$ (Alg.~\ref{alg:rapid}, line~\ref{ln:rrcp:while}).
At each iteration, the method samples a pickup station $i\in\mathcal{A}^{-}$ using Eq.~\eqref{eq:rapid_pick_prob} (Alg.~\ref{alg:rapid}, line~\ref{ln:rrcp:pick}).
Given $i$, the method builds $\mathrm{RCL}(i)\subseteq\mathcal{A}^{+}$ of size $m=\min\{m_{\max},|\mathcal{A}^{+}|\}$ using Eq.~\eqref{eq:rapid_rcl} (Alg.~\ref{alg:rapid}, line~\ref{ln:rrcp:rcl}).
The method computes $\mathrm{score}(i,j)$ for each $j\in\mathrm{RCL}(i)$ using one of Eqs.~\eqref{eq:rapid_score_svc}--\eqref{eq:rapid_score_spd} (Alg.~\ref{alg:rapid}, line~\ref{ln:rrcp:score}).
The method samples the drop station $j\in\mathrm{RCL}(i)$ using Eq.~\eqref{eq:rapid_drop_prob}.
The sampled pair $(i,j)$ is appended to a pair list $\mathcal{P}$, $\mathcal{P}\leftarrow \mathcal{P}\cup\{(i,j,q(i,j))\}$ and removed from availability, i.e., $\mathcal{A}^{-}\leftarrow \mathcal{A}^{-}\setminus\{i\}$ and $\mathcal{A}^{+}\leftarrow \mathcal{A}^{+}\setminus\{j\}$.
This update ensures that each station appears in at most one pair, which implies the single-visit constraints after routing.

\begin{equation} 
\Pr(i) \;=\; \frac{(\mathrm{sur}_i+\varepsilon)^{\beta_{\mathrm{pick}}}}
{\sum\limits_{u\in\mathcal{A}^{-}}(\mathrm{sur}_u+\varepsilon)^{\beta_{\mathrm{pick}}}},
\qquad \beta_{\mathrm{pick}}\ge 0,
\label{eq:rapid_pick_prob}
\end{equation}
\begin{equation}
\mathrm{RCL}(i)\subseteq \mathcal{A}^{+},\qquad |\mathrm{RCL}(i)|=m,
\label{eq:rapid_rcl}
\end{equation}
\begin{align}
\text{(A) service-only:}\quad & \mathrm{score}(i,j)=q(i,j), \label{eq:rapid_score_svc}\\
\text{(B) inverse-distance:}\quad & \mathrm{score}(i,j)=\frac{1}{d_{ij}+\varepsilon}, \label{eq:rapid_score_inv}\\
\text{(C) service-per-distance:}\quad & \mathrm{score}(i,j)=\frac{q(i,j)}{d_{ij}+\varepsilon}, \label{eq:rapid_score_spd}
\end{align}

\begin{equation}
\Pr(j\mid i) \;=\;
\frac{(\mathrm{score}(i,j)+\varepsilon)^{\beta_{\mathrm{drop}}}}
{\sum\limits_{v\in \mathrm{RCL}(i)}(\mathrm{score}(i,v)+\varepsilon)^{\beta_{\mathrm{drop}}}},
\qquad \beta_{\mathrm{drop}}\ge 0.
\label{eq:rapid_drop_prob}
\end{equation}


\paragraph{Stage 2: balanced insertion of pairs into truck routes.}
RRCP-BI initializes empty routes and per-truck state variables (Alg.~\ref{alg:rapid}, line~\ref{ln:rrcp:init_routes}).
RRCP-BI then processes each pair $(i,j)\in\mathcal{P}$ (Alg.~\ref{alg:rapid}, line~\ref{ln:rrcp:forpairs}) and assigns the pair to the truck that minimizes $\Delta_t(i,j)+\lambda n_t$ using Eqs.~\eqref{eq:rapid_incremental}--\eqref{eq:rapid_balanced_cost} (Alg.~\ref{alg:rapid}, line~\ref{ln:rrcp:assign}).
After selecting $t^\star$, the method appends $i$ and $j$ to $R_{t^\star}$ and updates $\ell_{t^\star}$ and $n_{t^\star}$ (Alg.~\ref{alg:rapid}, line~\ref{ln:rrcp:update}).
RRCP-BI returns the constructed route set $\mathbf{R}$ (Alg.~\ref{alg:rapid}, line~\ref{ln:rrcp:return}).


\begin{equation}
\Delta_t(i,j)=d_{\ell_t,i}+d_{i,j}+d_{j,0}-d_{\ell_t,0},
\label{eq:rapid_incremental}
\end{equation}

\begin{equation}
\mathrm{cost}_t(i,j)=\Delta_t(i,j)+\lambda n_t,\qquad \lambda\ge 0.
\label{eq:rapid_balanced_cost}
\end{equation}


\begin{algorithm}[h]
\caption{RRCP-BI}\label{alg:rapid}
\begin{algorithmic}[1]
\Require $(\mathcal{S},\mathcal{T},d_{ij},C,O_i)$; expected-demand proxy; $m_{\max}$; $\beta_{\mathrm{pick}},\beta_{\mathrm{drop}}$; score mode; $\lambda$; $\varepsilon>0$
\Ensure Route set $\mathbf{R}=\{R_t\}_{t\in\mathcal{T}}$
\State Compute $\widehat{D}_i$, $\mathrm{sur}_i$, $\mathrm{def}_i$ via Eqs.~\eqref{eq:rapid_proxy_need}--\eqref{eq:rapid_surdef} \label{ln:rrcp:proxy}
\State $\mathcal{A}^{-}\leftarrow\{i:\mathrm{sur}_i>0\}$; $\mathcal{A}^{+}\leftarrow\{i:\mathrm{def}_i>0\}$; $\mathcal{P}\leftarrow\emptyset$ \label{ln:rrcp:init_sets}
\While{$\mathcal{A}^{-}\neq\emptyset$ \textbf{and} $\mathcal{A}^{+}\neq\emptyset$} \label{ln:rrcp:while}
  \State Sample $i\in\mathcal{A}^{-}$ using Eq.~\eqref{eq:rapid_pick_prob} \label{ln:rrcp:pick}
  \State Build $\mathrm{RCL}(i)$ of size $m$ using Eq.~\eqref{eq:rapid_rcl} \label{ln:rrcp:rcl}
  \State Compute $\mathrm{score}(i,j)$ for $j\in\mathrm{RCL}(i)$ using Eqs.~\eqref{eq:rapid_score_svc}--\eqref{eq:rapid_score_spd} \label{ln:rrcp:score}
  \State $\mathcal{P}\leftarrow \mathcal{P}\cup\{(i,j,q(i,j))\}$
  \State $\mathcal{A}^{-}\leftarrow \mathcal{A}^{-}\setminus\{i\}$
  \State $\mathcal{A}^{+}\leftarrow \mathcal{A}^{+}\setminus\{j\}$
\EndWhile \label{ln:endwhile1}
\State Initialize $R_t\leftarrow()$, $\ell_t\leftarrow 0$, $n_t\leftarrow 0$ for all $t\in\mathcal{T}$ \label{ln:rrcp:init_routes}
\ForAll{$(i,j)\in\mathcal{P}$} \label{ln:rrcp:forpairs}
  \State $t^\star\leftarrow \arg\min_{t\in\mathcal{T}}\{\Delta_t(i,j)+\lambda n_t\}$ using Eqs.~\eqref{eq:rapid_incremental}--\eqref{eq:rapid_balanced_cost} \label{ln:rrcp:assign}
  \State Append $i,j$ to $R_{t^\star}$; update $\ell_{t^\star}\leftarrow j$; $n_{t^\star}\leftarrow n_{t^\star}+1$ \label{ln:rrcp:update}
\EndFor \label{ln:endfor2}
\State \Return $\mathbf{R}$ \label{ln:rrcp:return}
\end{algorithmic}
\end{algorithm}

\subsection{GLOBE: Global Score-Biased Pair-Move Greedy}\label{subsec:globe}

GLOBE is a greedy constructor that builds multi-truck routes by repeatedly selecting one pickup--drop pair-move across all trucks.
The method uses score-biased sampling over a global feasible move set and enforces the single-visit constraints by removing both selected stations after each move.
GLOBE uses the expected-demand proxy to guide construction, while objective values are computed by the deterministic evaluator in Section~\ref{sec:evaluation}.

GLOBE uses the proxy net desired change $\widehat{D}_i$ (Eq.~\eqref{eq:rapid_proxy_need}) and the corresponding $\mathrm{sur}_i$ and $\mathrm{def}_i$ magnitudes (Eq.~\eqref{eq:rapid_surdef}).
The method maintains a global visited set $U\subseteq\mathcal{S}$, per-truck current positions $u_t$, and per-truck routes $R_t$.
Initialization sets $U\leftarrow\varnothing$ and $u_t\leftarrow 0$ for all $t\in\mathcal{T}$ (Alg.~\ref{alg:globe}, \ref{ln:globe:init}).

Algorithm~\ref{alg:globe} summarizes GLOBE.
The method iterates by rebuilding the global feasible move set and sampling one move (Alg.~\ref{alg:globe}, \ref{ln:globe:while}--\ref{ln:globe:update}) until no feasible move remains (Alg.~\ref{alg:globe}, \ref{ln:globe:stopSD}--\ref{ln:globe:stopOmega}).

\paragraph{Parameters.}
GLOBE uses five user parameters.
$d_1>0$ bounds the distance from the current truck position to a pickup station and $d_2>0$ bounds the distance from pickup to drop.
$\gamma\ge 0$ controls selection pressure in the score-biased sampling distribution (Eq.~\eqref{eq:globe_prob}).
The \emph{score mode} selects the definition of $\mathrm{score}(t,i,j)$ (Eqs.~\eqref{eq:globe_score_serv}--\eqref{eq:globe_score_spd}).
$\varepsilon>0$ is a numerical-stability constant used in inverse-distance terms and in the sampling distribution.

\paragraph{Global feasible moves under distance thresholds.}
At each iteration, GLOBE defines $S$ and $D$, the available surplus and deficit sets among unvisited stations, respectively according to Eq.~\eqref{eq:globe_SD} (Alg.~\ref{alg:globe}, \ref{ln:globe:SD}).
A feasible pair-move is a triple $(t,i,j)$ with $t\in\mathcal{T}$, $i\in S$, and $j\in D$ that satisfies the distance thresholds $d(u_t,i)\le d_1$ and $d(i,j)\le d_2$.

\begin{equation}
S=\{i\in\mathcal{S}\setminus U:\mathrm{sur}_i>0\},\qquad
D=\{j\in\mathcal{S}\setminus U:\mathrm{def}_j>0\}.
\label{eq:globe_SD}
\end{equation}

For each feasible move, GLOBE defines the capacity-capped proxy service $q(i,j)$ and the incremental move distance $\Delta(t,i,j)$ according to Eq~\eqref{eq:globe_q_delta} (Alg.~\ref{alg:globe}, \ref{ln:globe:qdelta}).
GLOBE collects all feasible moves into a global candidate set $\Omega$ (Eq.~\eqref{eq:globe_Omega}). 
\begin{equation}
q(i,j)=\min\{\mathrm{sur}_i,\mathrm{def}_j,C\},\qquad
\Delta(t,i,j)=d(u_t,i)+d(i,j).
\label{eq:globe_q_delta}
\end{equation}
\begin{equation}
\Omega=\{(t,i,j): t\in\mathcal{T},\, i\in S,\, j\in D,\, d(u_t,i)\le d_1,\, d(i,j)\le d_2\}.
\label{eq:globe_Omega}
\end{equation}

For each $(t,i,j)\in\Omega$, GLOBE computes a nonnegative score $\mathrm{score}(t,i,j)$ using one of the score modes: service-only, inverse-distance, and service-per-distance (Eq.~\eqref{eq:globe_score_serv}--\eqref{eq:globe_score_spd}).

\begin{align}
\text{(A) service-only:}\quad & \mathrm{score}(t,i,j) = q(i,j), \label{eq:globe_score_serv}\\
\text{(B) inverse-distance:}\quad & \mathrm{score}(t,i,j)= \frac{1}{\Delta(t,i,j)+\varepsilon}, \label{eq:globe_score_dist} \\
\text{(C) service-per-distance:}\quad & \mathrm{score}(t,i,j)= \frac{q(i,j)}{\Delta(t,i,j)+\varepsilon}, \label{eq:globe_score_spd}
\end{align}

The method samples one move from $\Omega$ using score-biased probabilities $\Pr(t,i,j)$ (Eq.~\eqref{eq:globe_prob}).
After sampling $(t^\star,i^\star,j^\star)$, GLOBE appends $[i^\star,j^\star]$ to $R_{t^\star}$, updates $u_{t^\star}\leftarrow j^\star$, increments $m_{t^\star}$, and marks both stations as visited (Alg.~\ref{alg:globe}, lines~\ref{ln:globe:append}--\ref{ln:globe:update}).

\begin{equation}
\Pr(t,i,j) \;=\; \frac{(\mathrm{score}(t,i,j)+\varepsilon)^{\gamma}}
{\sum_{(t',i',j')\in\Omega}(\mathrm{score}'(t',i',j')+\varepsilon)^{\gamma}},
\qquad \gamma\ge 0.
\label{eq:globe_prob}
\end{equation}

The method terminates when $S=\emptyset$ or $D=\emptyset$, or when $\Omega=\varnothing$ (Alg.~\ref{alg:globe}, lines~\ref{ln:globe:stopSD}--\ref{ln:globe:stopOmega}).
If termination occurs with unvisited stations, the returned solution remains feasible under the single-visit constraints but can be partial.


\begin{algorithm}[h]
\caption{GLOBE}\label{alg:globe}
\begin{algorithmic}[1]
\Require $(\mathcal{S},\mathcal{T},d_{ij},C,O_i)$; expected-demand proxy; $d_1,d_2$; $\gamma$; $\lambda$; score mode; $\varepsilon>0$
\Ensure Route set $\mathbf{R}=\{R_t\}_{t\in\mathcal{T}}$
\State Compute $\widehat{D}_i$, $\mathrm{sur}_i$, $\mathrm{def}_i$ via Eqs.~\eqref{eq:rapid_proxy_need}--\eqref{eq:rapid_surdef} \label{ln:globe:proxy}
\State Initialize $R_t\leftarrow()$, $u_t\leftarrow 0$, $m_t\leftarrow 0$ for all $t\in\mathcal{T}$; $U\leftarrow\emptyset$ \label{ln:globe:init}
\State Build $S,D$ via Eq.~\eqref{eq:globe_SD} \label{ln:globe:SD}
\State Build $\Omega$ via Eq.~\eqref{eq:globe_Omega} \label{ln:globe:Omega_init}
\While{$S\neq\emptyset$ \textbf{and} $D\neq\emptyset$ \textbf{and} $\Omega\neq\emptyset$} \label{ln:globe:while}
  \State For all $(t,i,j)\in\Omega$, compute $q(i,j)$ and $\Delta(t,i,j)$ via Eq.~\eqref{eq:globe_q_delta} \label{ln:globe:qdelta}
  \State For all $(t,i,j)\in\Omega$, compute $\mathrm{score}(t,i,j)$ via Eqs.~\eqref{eq:globe_score_serv}--\eqref{eq:globe_score_spd}\label{ln:globe:score}
  \State Sample $(t^\star,i^\star,j^\star)\in\Omega$ via Eq.~\eqref{eq:globe_prob} \label{ln:globe:sample}
  \State $R_{t^\star} \leftarrow R_{t^\star}\,\|\, [i^\star,j^\star]$ \label{ln:globe:append}
  \State $u_{t^\star}\leftarrow j^\star$; $m_{t^\star}\leftarrow m_{t^\star}+1$; $U\leftarrow U\cup\{i^\star,j^\star\}$ \label{ln:globe:update}
  \State Build $S,D$ via Eq.~\eqref{eq:globe_SD} \label{ln:globe:stopSD}
  \State Build $\Omega$ via Eq.~\eqref{eq:globe_Omega} \label{ln:globe:stopOmega}

\EndWhile \label{ln:globe:endwhile}
\State \Return $\mathbf{R}$ \label{ln:globe:return}
\end{algorithmic}
\end{algorithm}

\section{Statistical Test}\label{app:statistics}
\begin{figure}[!ht]
    \centering
    \includegraphics[width=0.5\linewidth]{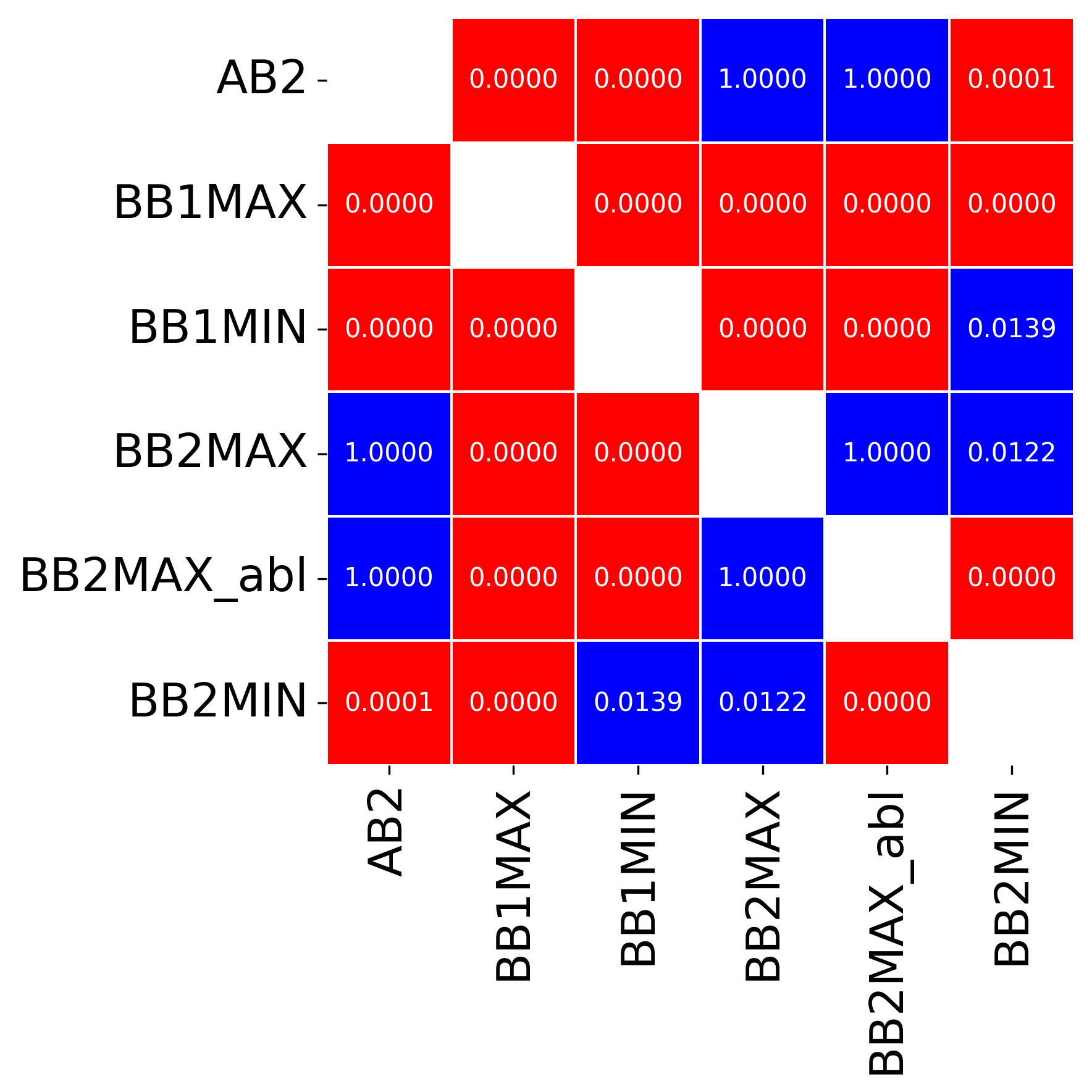}
    \caption{Heatmap of pairwise comparisons based on the rhv metric.}
    \label{heatmap:rhv}
\end{figure}

\begin{figure}
    \centering
    \includegraphics[width=0.5\linewidth]{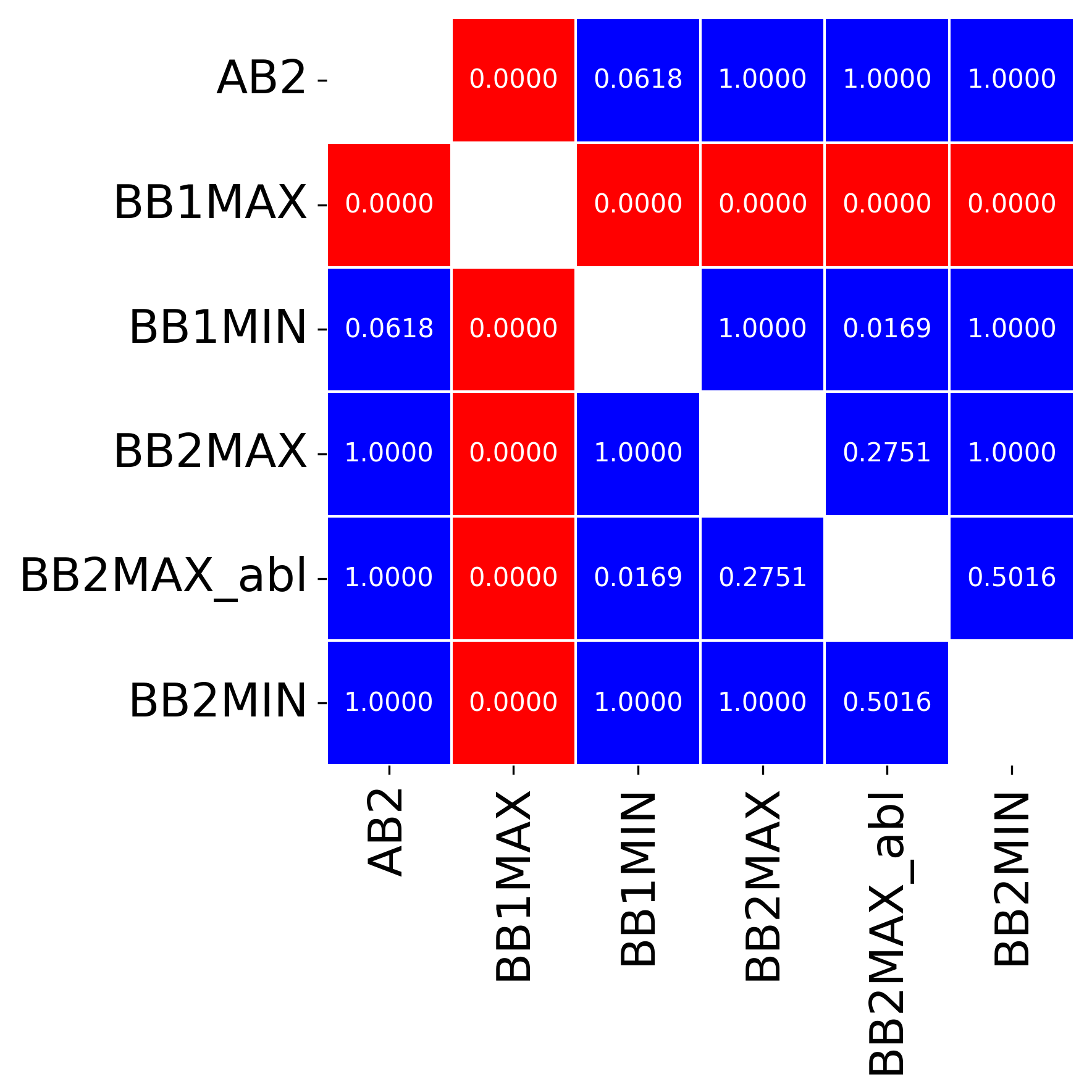}
    \caption{Heatmap of pairwise comparisons based on the gd+ metric. }
    \label{heatmap:gd}
\end{figure}

\begin{figure}
    \centering
    \includegraphics[width=0.5\linewidth]{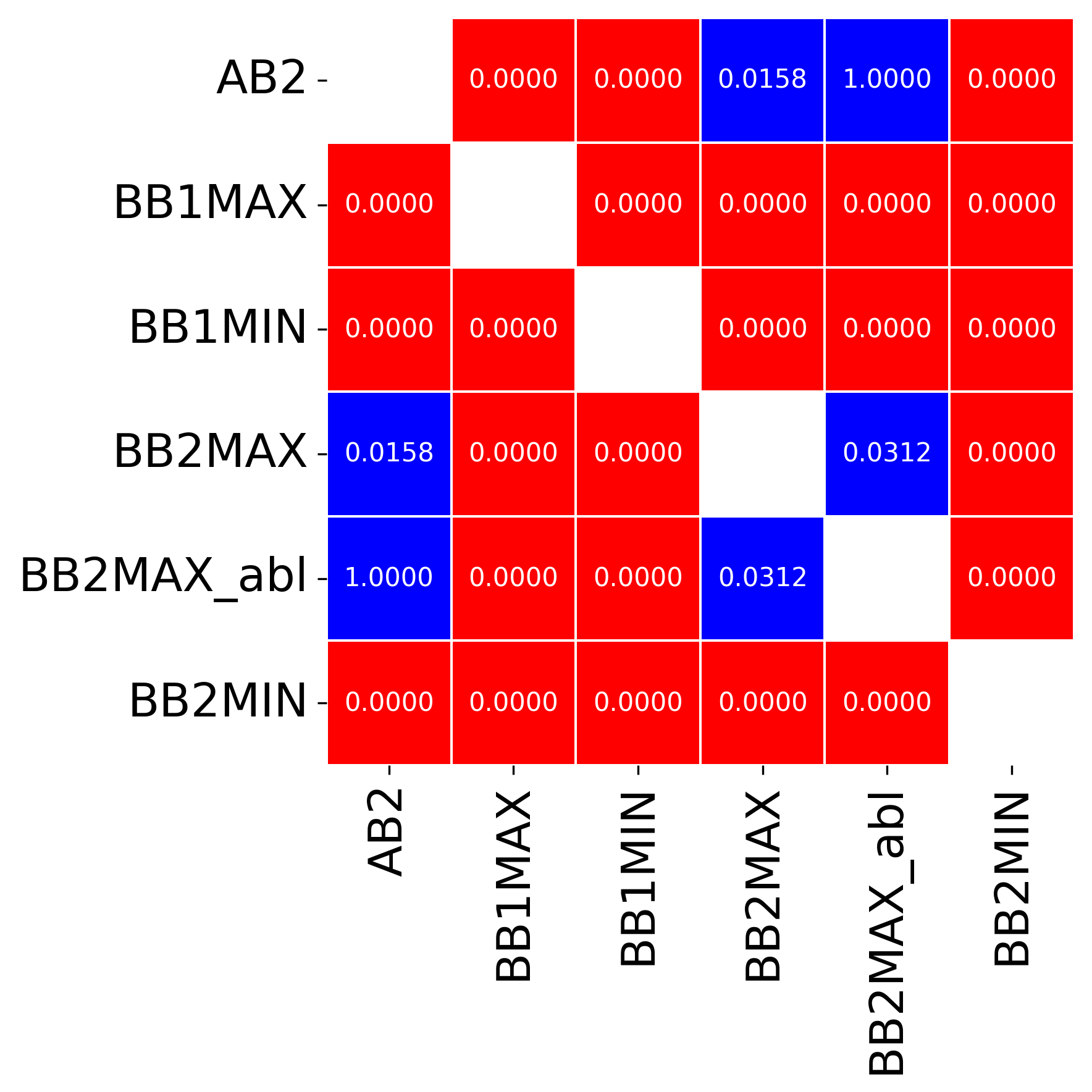}
    \caption{Heatmap of pairwise comparisons based on the igd+ metric. }
    \label{heatmap:igd}
\end{figure}

\begin{figure}
    \centering
    \includegraphics[width=0.5\linewidth]{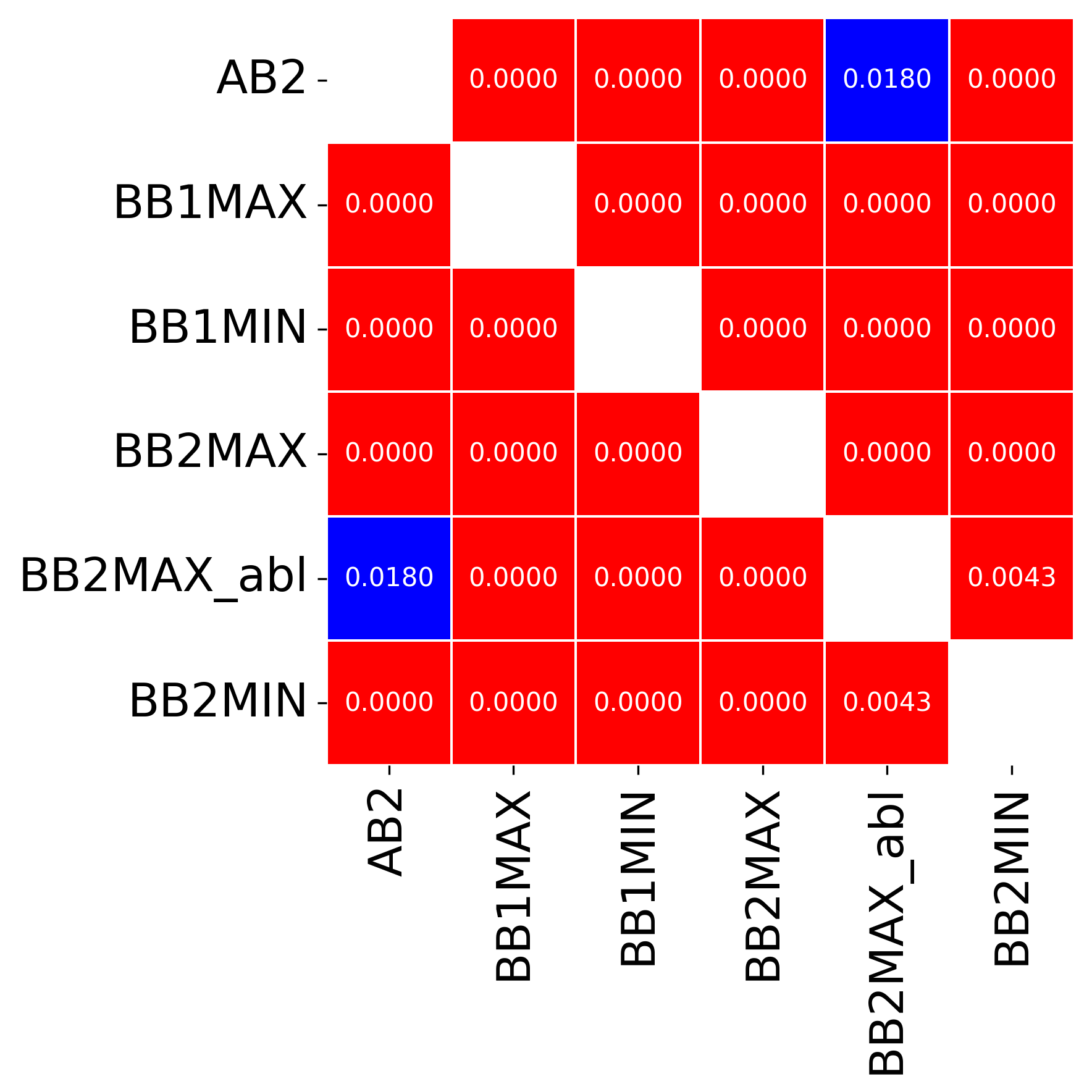}
    \caption{Heatmap of pairwise comparisons based on the spread metric.}
    \label{heatmap:spread}
\end{figure}

\begin{figure}
    \centering
    \includegraphics[width=0.5\linewidth]{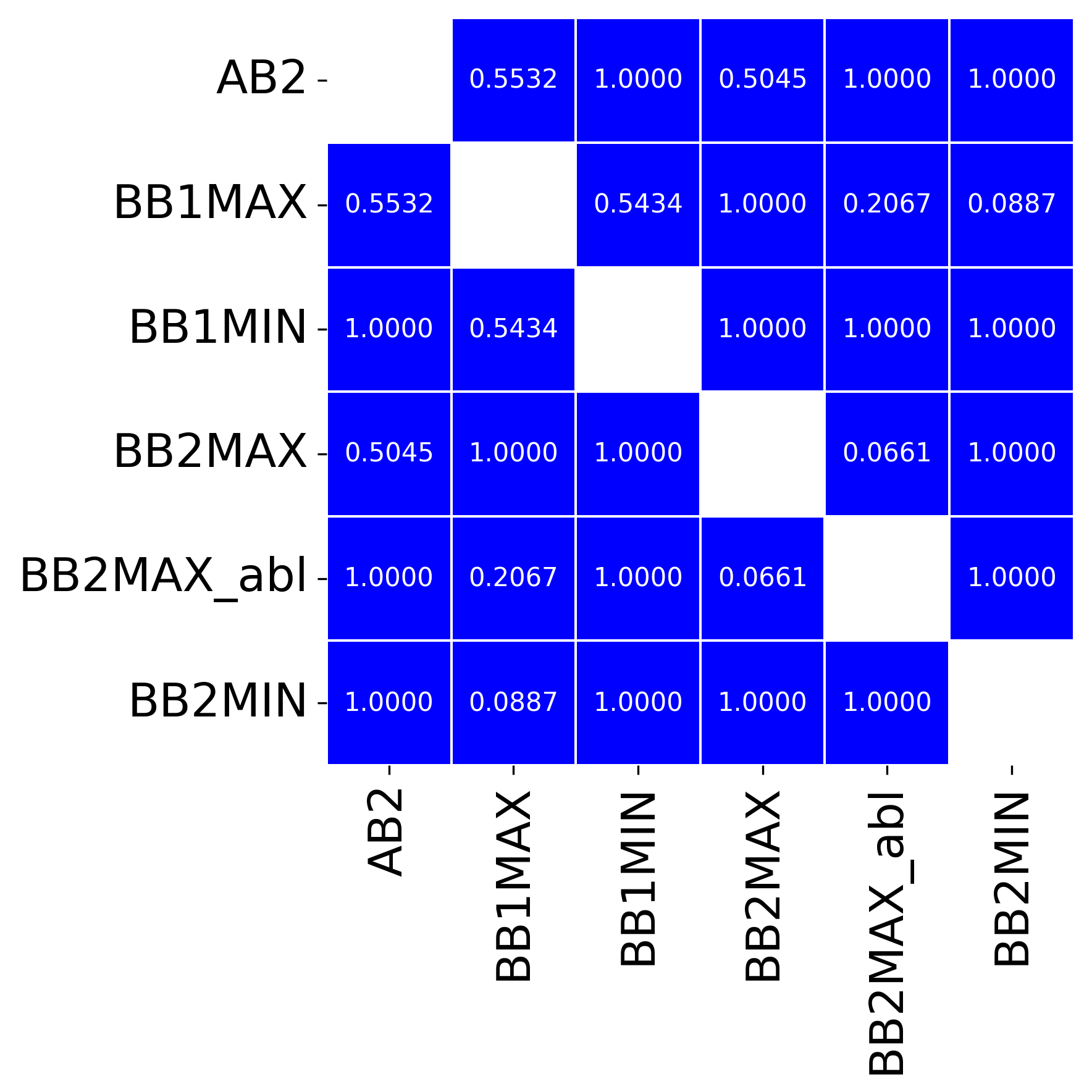}
    \caption{Heatmap of pairwise comparisons based on the \#nds metric.}
    \label{heatmap:ns}
\end{figure}

This section shows the statistical test of the mutation operators and the BB1-MAX ablated version. Figures~\ref{heatmap:rhv}-\ref{heatmap:ns} shows heatmaps of the pairwise comparison using Wilcoxon signed-rank test with Bonferroni ($\alpha < 0.01$). Red cells indicate significant differences, while blue cells indicate non-significant results.

\clearpage
\end{document}

%% file: problem_definition_jamal.tex
\section{Problem Description}\label{sec:problem}

This work considers a static (single-period) overnight rebalancing plan for a docked bike-sharing system.
A fleet of identical trucks departs from a single depot, visits a subset of stations, and returns to the depot within the period.
Each station is visited at most once and served by at most one truck.
Decision variables define truck routes, while station-level pickup and delivery quantities are computed by a deterministic recourse policy during scenario evaluation.

\subsection{Sets and parameters}
Let $\mathcal{T}$ denote the set of trucks, with $|\mathcal{T}|=T$.
Let $\mathcal{S}$ denote the set of stations, with $|\mathcal{S}|=S$, and let node $0$ denote the depot.
The set of all locations is $\mathcal{V}=\mathcal{S}\cup\{0\}$.
For any $i,j\in\mathcal{V}$, parameter $d_{ij}\ge 0$ denotes the travel distance from $i$ to $j$.
Each truck has capacity $C$ (bikes).
Each station $i\in\mathcal{S}$ has docking capacity $L_i$ (bikes) and initial inventory $O_i$ at the start of the planning period, with $0\le O_i\le L_i$.

Demand uncertainty is represented by a finite scenario set $\mathcal{H}$, with $|\mathcal{H}|=N$.
Parameter $D_{i,h}$ denotes the net desired inventory change at station $i\in\mathcal{S}$ under scenario $h\in\mathcal{H}$.
A positive value $D_{i,h}>0$ indicates a desired delivery of $D_{i,h}$ bikes to station $i$, whereas a negative value $D_{i,h}<0$ indicates a desired pickup of $|D_{i,h}|$ bikes from station~$i$.
Parameter $p_h$ denotes the nonnegative weight of scenario $h\in\mathcal{H}$, with $\sum_{h\in\mathcal{H}} p_h = 1$.
When calibrated probabilities are unavailable, uniform weights $p_h = 1/|\mathcal{H}|$ is adopted.

A target inventory $\tau_{i,h}$ (Eq.~\eqref{eq:target_inventory}) is obtained by projecting the requested inventory $O_i + D_{i,h}$ into the feasible station range:
\begin{equation}
\tau_{i,h} \;=\; \min\{L_i,\max\{0,\, O_i + D_{i,h}\}\},
\qquad \forall i\in\mathcal{S},\; h\in\mathcal{H}.
\label{eq:target_inventory}
\end{equation}

This work defines the \emph{demand intensity} of a scenario as the total magnitude of requested inventory changes:
\begin{equation}
\Phi_h \;=\; \sum_{i\in\mathcal{S}} |D_{i,h}|,
\qquad \forall h\in\mathcal{H}.
\label{eq:scenario_intensity}
\end{equation}
Let $\kappa_{0.90}$ denote the empirical $90$th percentile of $\{\Phi_h\}_{h\in\mathcal{H}}$.
The subset of high-demand scenarios is
\begin{equation}
\mathcal{H}^{0.90} \;=\; \{h\in\mathcal{H}\,:\,\Phi_h \ge \kappa_{0.90}\}.
\label{eq:high_demand_set}
\end{equation}
For aggregation within $\mathcal{H}^{0.90}$, this work uses normalized weights
\begin{equation}
\tilde{p}_h \;=\; \frac{p_h}{\sum_{h'\in\mathcal{H}^{0.90}} p_{h'}},
\qquad \forall h\in\mathcal{H}^{0.90}.
\label{eq:normalized_high_demand_weights}
\end{equation}

\subsection{Representation and route feasibility}
A solution is represented by a set of truck routes $\mathbf{R}=\{R_t\}_{t\in\mathcal{T}}$.
Each route $R_t$ is an ordered sequence of stations assigned to truck $t$.
Eq.~\eqref{eq:route_def} defines the representation, where $k_t$ denotes the number of stations visited by truck $t$.
A non-empty route starts at the depot, visits stations in order, and returns to the depot.
The empty route (i.e., $k_t=0$) represents an unused truck.

\begin{equation}
R_t=(r_{t,1}, r_{t,2}, \ldots, r_{t,k_t}),
\qquad r_{t,j}\in\mathcal{S},\quad k_t\ge 0.
\label{eq:route_def}
\end{equation}

Two feasibility restrictions are imposed on the route structure.
Eq.~\eqref{eq:unique_service} enforces that each station is served by at most one truck.
Eq.~\eqref{eq:no_repeat_within_route} enforces that a truck does not visit the same station more than once.
The indicator function $\mathbb{I}(\cdot)$ takes value $1$ if its argument is true and value $0$ otherwise.
\begin{equation}
\sum_{t\in\mathcal{T}} \sum_{j=1}^{k_t} \mathbb{I}(r_{t,j}=i) \;\le\; 1,
\qquad \forall i\in\mathcal{S}.
\label{eq:unique_service}
\end{equation}
\begin{equation}
\mathbb{I}(r_{t,j}=r_{t,\ell}) = 0,
\qquad \forall t\in\mathcal{T},\;\forall j\neq \ell,\; j,\ell\in\{1,\ldots,k_t\}.
\label{eq:no_repeat_within_route}
\end{equation}

\subsection{Objective functions}
The optimization problem is tri-objective.
The first objective minimizes total travel distance by the truck fleet (Eq.~\eqref{eq:obj_distance}).
The second objective minimizes scenario-weighted unmet demand over all scenarios (Eq.~\eqref{eq:obj_unmet}).
The third objective minimizes unmet demand under high-demand scenarios (Eq.~\eqref{eq:obj_unmet_p90}), which explicitly targets robustness in the upper tail of scenario demand intensity.

\begin{equation}
f_1(\mathbf{R}) \;=\; f_{\mathrm{dist}}(\mathbf{R})
\;=\;\sum_{t\in\mathcal{T}} \delta(R_t).
\label{eq:obj_distance}
\end{equation}

\begin{equation}
\delta(R_t)=
\begin{cases}
d_{0,r_{t,1}}+\sum_{j=1}^{k_t-1} d_{r_{t,j},r_{t,j+1}} + d_{r_{t,k_t},0}, & k_t\ge 1,\\[2mm]
0, & k_t=0.
\end{cases}
\label{eq:route_distance}
\end{equation}

For each scenario $h\in\mathcal{H}$, the evaluation procedure computes the station-level unmet demand $U_{i,h}(\mathbf{R})$ after executing the route set $\mathbf{R}$ under the deterministic recourse policy in Section~\ref{sec:evaluation}.
The scenario-level unmet demand is $U_h(\mathbf{R})=\sum_{i\in\mathcal{S}} U_{i,h}(\mathbf{R})$.
The scenario-weighted unmet demand objective is
\begin{equation}
f_2(\mathbf{R})=f_{\mathrm{ud}}(\mathbf{R})
=\sum_{h\in\mathcal{H}} p_h \, U_h(\mathbf{R})
=\sum_{h\in\mathcal{H}} p_h \sum_{i\in\mathcal{S}} U_{i,h}(\mathbf{R}).
\label{eq:obj_unmet}
\end{equation}

The high-demand robustness objective aggregates unmet demand only over $\mathcal{H}^{0.90}$:
\begin{equation}
f_3(\mathbf{R})=f_{\mathrm{ud}}^{0.90}(\mathbf{R})
=\sum_{h\in\mathcal{H}^{0.90}} \tilde{p}_h \, U_h(\mathbf{R})
=\sum_{h\in\mathcal{H}^{0.90}} \tilde{p}_h \sum_{i\in\mathcal{S}} U_{i,h}(\mathbf{R}).
\label{eq:obj_unmet_p90}
\end{equation}

Eq.~\eqref{eq:triobj_problem} states the tri-objective rebalancing problem.
The solution set consists of Pareto-non-dominated feasible route sets.
\begin{equation}
\min_{\mathbf{R}} \;\; \Big(f_1(\mathbf{R}),\, f_2(\mathbf{R}),\, f_3(\mathbf{R})\Big)
\quad \text{subject to Eqs.~\eqref{eq:route_def}--\eqref{eq:no_repeat_within_route}.}
\label{eq:triobj_problem}
\end{equation}

\subsection{Scenario evaluation under a deterministic recourse policy} \label{sec:evaluation}
A route set $\mathbf{R}$ is evaluated on each scenario $h\in\mathcal{H}$ through a simulation that enforces truck and station capacity constraints, obtaining as resutl the unmet demand values $U_{i,h}(\mathbf{R})$ for all stations.

Variable $s_{i,h}$ denotes the station inventory at station $i\in\mathcal{S}$ during the simulation.
Variable $q_{t,h}$ denotes the truck load of truck $t\in\mathcal{T}$ during the simulation.
Eq.~\eqref{eq:init_states} defines the initial state.
At a visit of truck $t\in\mathcal{T}$ to station $i=r_{t,j}$, the desired inventory change is defined by Eq.~\eqref{eq:desired_change}.

\begin{equation}
s_{i,h}\leftarrow O_i,\;\;\forall i\in\mathcal{S},
\qquad
q_{t,h}\leftarrow 0,\;\;\forall t\in\mathcal{T}.
\label{eq:init_states}
\end{equation}
\begin{equation}
\Delta_{i,h} \;=\; \tau_{i,h} - s_{i,h}.
\label{eq:desired_change}
\end{equation}

Variable $y_{t,i,h}$ denotes the executed transfer at station $i$ by truck $t$ under scenario $h$.
A positive value of $y_{t,i,h}$ denotes a delivery to the station, and a negative value of $y_{t,i,h}$ indicates a pickup from the station.
Eq.~\eqref{eq:feasible_transfer_interval} defines the feasible transfer interval induced by truck capacity and station capacity.
\begin{equation}
-\min\{C-q_{t,h},\, s_{i,h}\}
\;\le\;
y_{t,i,h}
\;\le\;
\min\{q_{t,h},\, L_i-s_{i,h}\}.
\label{eq:feasible_transfer_interval}
\end{equation}

The recourse policy applies a projection rule that selects the feasible transfer closest to the desired change.
Eq.~\eqref{eq:projection_rule} defines the rule, where $\Pi_{[a,b]}(x)=\min\{b,\max\{a,x\}\}$ and where $[\underline{y}_{t,i,h},\overline{y}_{t,i,h}]$ denotes the interval in Eq.~\eqref{eq:feasible_transfer_interval}.
\begin{equation}
y_{t,i,h} \;=\; \Pi_{[\underline{y}_{t,i,h},\,\overline{y}_{t,i,h}]}\!\left(\Delta_{i,h}\right).
\label{eq:projection_rule}
\end{equation}

Eq.~\eqref{eq:state_updates} defines the state updates after executing the transfer.
Eq.~\eqref{eq:unmet_definition} defines the unmet demand at a visited station as the residual between the desired change and the executed transfer.
If station $i\in\mathcal{S}$ is not visited by any truck in $\mathbf{R}$, then the station inventory remains equal to $O_i$, i.e., $U_{i,h}(\mathbf{R}) \;=\; \left|\tau_{i,h} - O_i\right|$.
\begin{equation}
s_{i,h} \leftarrow s_{i,h} + y_{t,i,h},
\qquad
q_{t,h} \leftarrow q_{t,h} - y_{t,i,h}.
\label{eq:state_updates}
\end{equation}
\begin{equation}
U_{i,h}(\mathbf{R}) \leftarrow \left|\Delta_{i,h} - y_{t,i,h}\right|.
\label{eq:unmet_definition}
\end{equation}